\documentclass{article} 
\usepackage{iclr2026_conference,times}

\usepackage{amsmath,amsfonts,bm}

\def\eqref#1{equation~\ref{#1}}

\def\1{\bm{1}}

\DeclareMathAlphabet{\mathsfit}{\encodingdefault}{\sfdefault}{m}{sl}
\SetMathAlphabet{\mathsfit}{bold}{\encodingdefault}{\sfdefault}{bx}{n}

\usepackage{hyperref}
\usepackage{url}

\usepackage{booktabs} 

\usepackage{makecell}

\usepackage{float}

\usepackage{enumitem}

\makeatletter
\renewcommand{\@fnsymbol}[1]{}
\makeatother

\hypersetup{hidelinks}

\usepackage[most]{tcolorbox}
\definecolor{cvteal}{RGB}{20,116,110}
\definecolor{cvback}{RGB}{248,253,251} 

\newtcolorbox{finding}[1]{%
  enhanced, breakable,
  colback=cvback,           
  colframe=cvteal!80,
  boxrule=0.6pt,
  arc=2mm,                  
  left=6pt,right=6pt,top=6pt,bottom=6pt,
  drop fuzzy shadow,        
  before upper={\textbf{Finding~#1:}\ } 
}

\title{A Comprehensive Information-Decomposition Analysis of Large Vision-Language Models}

\author{
  Lixin Xiu\textsuperscript{1}, Xufang Luo\textsuperscript{2}\thanks{Correspondence to: Xufang Luo \texttt{<xufluo@microsoft.com>} and Hideki Nakayama \texttt{<nakayama@ci.i.u-tokyo.ac.jp>}.}, Hideki Nakayama\textsuperscript{1} \\
  \textsuperscript{1}The University of Tokyo \\
  \textsuperscript{2}Microsoft Research
}

\iclrfinalcopy 
\begin{document}

\maketitle

\begin{abstract}
Large vision-language models (LVLMs) achieve impressive performance, yet their internal decision-making processes remain opaque, 
making it difficult to determine if the success stems from true multimodal fusion or from reliance on unimodal priors. To address this attribution gap, we introduce a novel 
framework using partial information decomposition (PID) to quantitatively 
measure the ``information spectrum'' of LVLMs---decomposing a model's decision-relevant information into redundant, unique, and synergistic components. 
By adapting a scalable estimator to modern LVLM outputs, our model-agnostic pipeline profiles 26 LVLMs on four datasets across three dimensions---\emph{breadth} (cross-model \& cross-task), \emph{depth} (layer-wise information dynamics), and \emph{time} (learning dynamics across training). 
Our analysis reveals two key results: (i) two task regimes (synergy-driven vs.\ knowledge-driven) and (ii) two stable, contrasting family-level strategies (fusion-centric vs.\ language-centric). We also uncover a consistent three-phase pattern in layer-wise processing and identify visual 
instruction tuning as the key stage where fusion is learned. Together, these contributions provide a quantitative lens beyond accuracy-only evaluation and offer insights for analyzing and designing the next generation of LVLMs.
Code and data are available at \url{https://github.com/RiiShin/pid-lvlm-analysis}.
\end{abstract}

\section{Introduction}

Large vision-language models (LVLMs) achieve remarkable success across 
a wide range of multimodal tasks, including 
visual question answering~\citep{chen2024internvl25}, image captioning~\citep{bai2025qwen25}, 
and open-ended reasoning~\citep{zhu2025internvl3}. However, the internal mechanisms driving 
these impressive results remain largely opaque. 
Accuracy and other aggregate performance metrics only 
reflect the final outcomes of model predictions, 
but not the underlying processes through which these outcomes are obtained. 
While prior research has begun to analyze 
large language models (LLMs) to isolate factors 
shaping predictions~\citep{jain2019attention,meng2022locating}, 
LVLMs pose a distinct challenge because they must process and integrate multiple modalities. 
In particular, understanding 
whether a model's prediction is primarily driven by visual evidence, 
language priors, or the interaction between the two is 
critical for interpreting its behavior.
However, existing interpretability efforts often adopt a ``micro-scope'' focus---analyzing one 
modality in isolation---or introduce ad hoc metrics that lack firm 
theoretical support. Consequently, the field 
still lacks comprehensive, quantitative tools capable of 
dissecting the internal strategies by which LVLMs use multimodal information.

To address this challenge, we introduce a framework built on partial 
information decomposition (PID) to comprehensively 
and quantitatively analyze LVLM behavior. PID is a rigorous information-theoretic 
methodology that decomposes the mutual information between a set of inputs and an output 
into distinct components. 
Originally developed in neuroscience and 
complex systems~\citep{williams2010nonnegative}, 
PID characterizes how multiple information channels jointly influence a target variable. 
We extend this perspective to LVLM inference 
by treating vision features $X_1$ and language features $X_2$ as 
inputs and the model prediction $Y$ as the output, 
partitioning decision-relevant information into four 
non-negative terms: redundancy $R$ (shared by both), vision uniqueness $U_1$, 
language uniqueness $U_2$, and synergy $S$ (emerging only from their combination).
We refer to $\{R,U_1,U_2,S\}$ as the model's \emph{information spectrum}.
It provides a principled lens for probing LVLM internals and, unlike ``micro-scope'' analyses or 
more empirical approaches, enables a quantitative separation of the model's core information-processing strategies.

\begin{figure}[t!]
    \centering
    \includegraphics[width=0.9\textwidth]{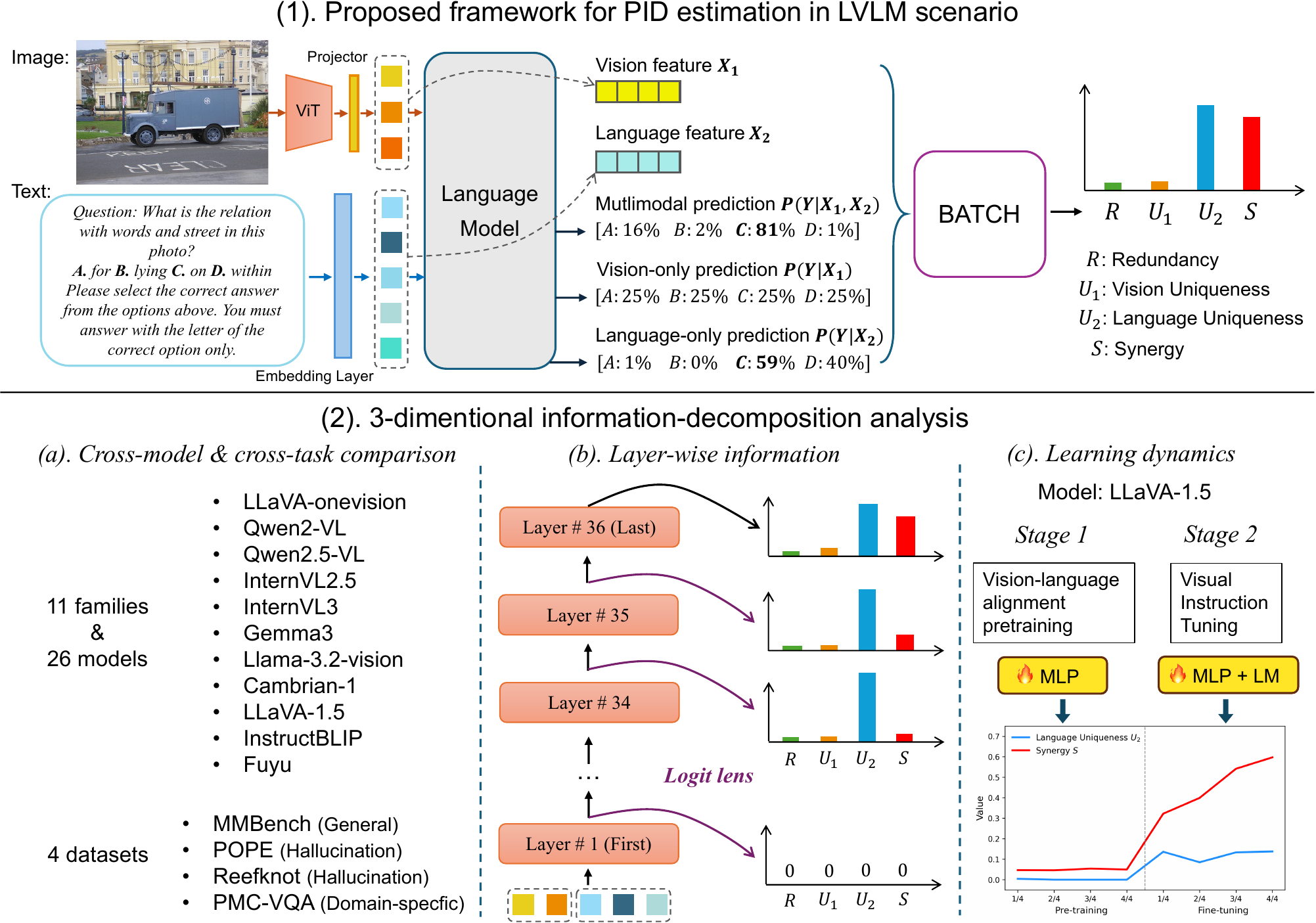}
    \caption{Overview of this research. The first part is the framework of PID estimation for LVLMs. 
    Given an image-text pair, we extract image and text embeddings as two features, run a standard multimodal forward pass
    and collect two unimodal predictions by masking the other modality. PID values are estimated with BATCH estimator.
    The second part reveals three analysis dimensions: (1) cross-model
    and cross-task comparison, (2) layer-wise information dynamics, and (3) learning dynamics over training. 
    To our knowledge, this is the first comprehensive LVLM analysis
    through the lens of information decomposition.}
    \label{fig:overview}
\end{figure}

To apply PID to modern LVLMs, we adapt the PID estimator proposed by \citet{liang2023} 
for visual question-answering (VQA) tasks and propose a model-agnostic pipeline that requires no architectural changes or retraining. 
Using this pipeline, we analyze LVLMs along three axes: (1) a cross-model, cross-task comparison spanning 
26 models and four benchmarks; (2) layer-wise information dynamics via a logit-lens view; 
and (3) the evolution of multimodal fusion across training stages. The full framework and 
analysis dimensions are summarized in Figure~\ref{fig:overview}.

Our study yields several actionable insights. We identify two task 
regimes---\emph{synergy-driven} vs.\ \emph{knowledge-driven}---and show that model families 
themselves adopt two stable, contrasting strategies---\emph{fusion-centric} vs.\ \emph{language-centric}. 
We clarify how fusion develops, observing a three-phase pattern over layers and 
finding that visual instruction tuning is the key stage where $S$ is unlocked. 
Taken together, these results provide a quantitative basis for moving beyond 
accuracy-only evaluation toward a more principled understanding of multimodal processing in LVLMs.

\section{Related work}
\label{rel_work}

\paragraph{The evolution of vision-language models.}
Vision-language models (VLMs) have shifted from contrastive dual-encoder paradigms 
to generative paradigms. 
While early dual-encoder models like CLIP~\citep{radford2021learning} and 
ALIGN~\citep{jia2021scaling} focused on joint representation 
learning, the current approach uses a 
parameter-efficient generative architecture, typically comprising a vision 
encoder, a projector, and a large language model (LLM) 
backbone~\citep{tsimpoukelli2021multimodal,alayrac2022flamingo,merullo2022linearly,li2023blip2}.

The evolution of large vision-language models (LVLMs) is driven by advances in LLM backbones and training 
methodologies~\citep{dai2024nvlm}. Architecturally, backbones have shifted from 
encoder-decoder models like T5~\citep{raffel2020t5} to decoder-only models such as Llama~\citep{touvron2023llama}. 
Methodologically, performance has been greatly enhanced by various training stages on vast, high-quality datasets, 
a strategy pioneered by LLaVA~\citep{liu2023llava} and 
MiniGPT-4~\citep{zhu2023minigpt}. State-of-the-art LVLMs integrate 
the most powerful backbones (e.g., Llama3.1~\citep{grattafiori2024llama} and Qwen2.5~\citep{qwen25lm}) with advanced training 
recipes~\citep{liu2024llava15,li2024llavaov,wang2024qwen2,chen2024internvl25,tong2024cambrian,meta2024llama31,bai2025qwen25,zhu2025internvl3}, 
resulting in more capable yet more complex systems.

\paragraph{Probing the black box: interpretability in VLMs.}
To address the opacity of VLMs, researchers have adapted interpretability 
techniques from transformer-based language models~\citep{vaswani2017attention}. 
One line of work generates post-hoc explanations to identify 
influential inputs using attribution 
heatmaps~\citep{schulz2020attribution, wang2023ib6}, 
attention maps~\citep{abnar2020tokenxai,chefer2021attention,clipxai1}, 
and activation analysis~\citep{conmy2023activation,arditi2024xmlnew}.
Complementary approaches probe a model's internal representations 
to understand its learned knowledge~\citep{meng2022locating}, for instance by 
training linear probes~\citep{alain2016prob1} to decode features in LLMs~\citep{hewitt2019prob2,tenney2019prob3}, 
applying the logit lens~\citep{nostalgebraist2020logitlens,tunedlens} to 
inspect LLMs' intermediate computations~\citep{cywinski2025logitlen2, neo2024logitlen3}, 
or identifying ``multimodal neurons'' corresponding to 
human-interpretable concepts~\citep{multineuro1, multineuro2}.

With the rise of LVLMs, research is shifting toward understanding 
their deep multimodal fusion. Recent work in this area includes 
analyzing and manipulating the visual token representations that 
bridge the two modalities~\citep{jiang2024xml1,basu2024xml3,liu2025xml2}, 
as well as quantifying vision's contribution through analyses of 
visual attention sinks~\citep{kang2025see} and cross-modal 
information-flow tracing~\citep{zhang2025cross,vlmflow2,yang2024law}.

\paragraph{An information-theoretic lens on multimodal learning.}
Information theory offers a quantitative framework for 
analyzing information flow in neural networks. Foundational 
concepts like mutual information (MI)~\citep{shannon1948mathematical} and 
the subsequent information bottleneck (IB) principle~\citep{tishby2000information} 
have been widely used in multimodal learning. 
Applications range from using MI for interpretability~\citep{oh2025mixai} to 
employing the IB framework for 
both guiding representation learning~\citep{ib1,jiang2024ib2,xiao2024ib3,wu2025ib4,bai2025ib7} 
and enhancing model transparency~\citep{wang2023ib6,zhu2025ib5}.

While mutual information can quantify the total information from a source, 
it cannot disentangle the complex interactions between multiple inputs, 
such as vision and text. To address this, partial information decomposition 
(PID)~\citep{williams2010nonnegative} decomposes the 
information about a target variable into redundant, unique, and synergistic components. 
The application of PID to machine learning is a 
nascent field~\citep{ehrlich2022pid1,dissanayake2025pid2,choi2025pid3,shan2025pid4}, 
though recent studies have begun to apply it 
within multimodal learning contexts~\citep{liang2023,liangpid2,liangpid3}. 
However, its use for analyzing the composition, flow, and evolution of 
multimodal information within modern LVLMs remains unexplored, which is a 
gap this paper aims to fill.

\section{Methodology}
\label{method}

This section details our methodology for applying PID to analyze LVLMs in three parts. 
We first review the fundamentals of PID and 
the specific estimator our work adapts in Section~\ref{sec:preliminaries}. 
We then present our framework, including three key adaptations that make this estimator 
robust for the unique context of modern LVLMs, in Section~\ref{sec:framework}. 
Finally, we outline the experimental design that leverages this framework to 
conduct a comprehensive analysis across three dimensions in Section~\ref{sec:exp_design}.

\subsection{Preliminaries}
\label{sec:preliminaries}

\paragraph{Partial information decomposition.}
Mutual information~\citep{shannon1948mathematical} measures the statistical dependence between two variables. 
However, in a three-variable system comprising two source variables $X_1, X_2$ and a target variable $Y$, 
with respective state spaces $\mathcal{X}_1, \mathcal{X}_2$, and $\mathcal{Y}$, 
the standard interaction information $I(X_1; X_2; Y)$ can be negative, 
limiting its interpretability.

To address this, partial information decomposition (PID)~\citep{williams2010nonnegative} reframes the problem 
by decomposing the total mutual information $I(X_1, X_2; Y)$ into 3 
non-negative atoms: \textbf{redundancy} (information common to both sources), 
\textbf{uniqueness} (information exclusive to each source), 
and \textbf{synergy} (new information emerging from their combination). 
Following \citet{bertschinger2014quantifying}, the components are 
defined on the set of 
distributions $\Delta_{P} = \bigl\{\, Q \in \Delta : Q(x_i,y) = P(x_i,y),\ \forall\, x_i \in \mathcal{X}_i,\ y \in \mathcal{Y},\ i \in \{1,2\} \,\bigr\}, $ which 
contains all joint distributions $Q$ over ($X_1$, $X_2$, $Y$) that preserve the source-target marginals of the true distribution $P$. The atoms are given as:
\begin{align}
R &= \max_{Q \in \Delta_P} I_Q(X_1; X_2; Y), \\
U_1 &= \min_{Q \in \Delta_P} I_Q(X_1; Y  \mid  X_2), \\
U_2 &= \min_{Q \in \Delta_P} I_Q(X_2; Y  \mid  X_1), \\
S &= I(X_1, X_2; Y) - \min_{Q \in \Delta_P} I_Q(X_1, X_2; Y).
\end{align}
This decomposition provides a principled framework for quantifying 
how individual and joint sources of information 
contribute to a target variable, 
but estimating these atoms from data is a non-trivial task that requires specialized estimators.

\paragraph{Estimating PID for multimodal inputs.}
To estimate PID for the high-dimensional representations within modern LVLMs, 
we adapt the scalable estimator from \citet{liang2023}. 
This work introduces two methods: 
a convex programming-based estimator CVX for discrete features, 
and an approximate estimator \textbf{BATCH} designed for 
continuous, high-dimensional modalities.

Our work builds upon the BATCH estimator, as it is well-suited for 
analyzing the continuous vectorial embeddings produced by LVLMs. 
This method uses neural networks to parameterize the required 
probability distributions. It then optimizes an information-theoretic 
objective over mini-batches, employing a variant of 
the Sinkhorn algorithm~\citep{cuturi2013sinkhorn} to 
enforce the marginal-matching constraints defined in $\Delta_P$.

See Appendix~\ref{appendix:pid} for more details on PID and the BATCH estimator.

\subsection{A PID Estimation Framework for LVLMs}
\label{sec:framework}

We propose a PID estimation framework for LVLMs tailored to 
multiple-choice visual question answering (MC-VQA) tasks. 
This focus is a deliberate methodological choice: BATCH requires a finite $\mathcal{Y}$, 
thus the natural set of choices in MC-VQA (e.g., $\{A, B, C, D\}$) 
allows for a clean analysis while avoiding the noisy and potentially 
biased process of manually clustering open-ended answers, or training 
auxiliary projection heads to map LVLM representations into pre-defined clusters. 
Such additional components make PID estimates sensitive to clustering and projection hyperparameters, making it unclear whether the estimated 
quantities primarily reflect the LVLM's original end-to-end behavior or 
the added mapping, which is not how these models are typically used.

As illustrated in Figure~\ref{fig:overview}~(1), 
our pipeline begins by defining the source 
variables---vision ($X_1$) and language ($X_2$)---from an LVLM's 
internal embeddings. 
We then conduct multimodal and unimodal inference runs to 
obtain three conditional probability 
distributions: $P(Y \mid X_1, X_2)$, $P(Y \mid X_1)$, and $P(Y \mid X_2)$. 
These distributions, along with the source features $X_1$ and $X_2$, 
are then fed into the BATCH estimator to compute 
the final PID values $\{R, U_1, U_2, S\}$. Notably, this estimation relies only on the 
model's predictive distributions and input
representations; no ground-truth labels are used when computing the PID components.
This makes PID a process-level descriptor of model behavior, complementary to standard accuracy metrics.


\paragraph{Input representation and unimodal conditioning.}
We define the source variables $X_1$ and $X_2$ as the mean-pooled visual and textual token embeddings,
respectively, and ablate alternative summarization 
strategies (last-hidden and max pooling) in Appendix~\ref{appendix:ablation}. 

Estimating the unimodal conditionals $P(Y \mid  X_1)$ and $P(Y \mid  X_2)$ for an integrated LVLM
requires a carefully designed probe. We approximate unimodal conditioning by masking one modality at the embedding level: 
following the corruption scheme of \citet{meng2022locating}, we replace the 
entire embedding sequence of the other modality with noise.
Each vector in this noise sequence is drawn i.i.d. from
$\mathcal{N}(\boldsymbol{\mu}, \operatorname{diag}(\boldsymbol{\sigma}^2))$,
where $\boldsymbol{\mu}, \boldsymbol{\sigma} \in \mathbb{R}^d $ denote the dimension-wise mean and standard deviation of that modality's embeddings, pre-computed across the dataset.
This calibrated noise removes the other modality while keeping the embedding scale in-distribution.

\paragraph{Confidence thresholding for renormalization.}
For both unimodal and multimodal conditioning, we extract a categorical predictive distribution 
over the finite candidate set $\mathcal{Y}$.
We first compute a token-length normalized candidate score
$S_{\text{orig}}(Y{=}y \mid \cdot)$ from the model log-likelihood of the candidate answer string,
and then renormalize across candidates to obtain $P(Y \mid \cdot)$.

However, renormalizing over a restricted candidate set can artificially inflate confidence when the model assigns low scores to all candidates under the full vocabulary distribution.
To mitigate overconfidence from a restricted candidate set,
we compute the total candidate-set score and apply a confidence threshold:
\begin{align}
\hat{P}(Y \mid \cdot)=
\begin{cases}
P(Y \mid \cdot) & \text{if } \sum_{y \in \mathcal{Y}} S_{\text{orig}}(Y{=}y \mid \cdot) \ge \tau \\
\mathcal{U}(K) & \text{otherwise}
\end{cases}
\end{align}
where $K=|\mathcal{Y}|$ and $\mathcal{U}(K)$ denotes the uniform distribution over $\mathcal{Y}$.
This prevents low-confidence guesses from contributing spurious structure to the PID computation.
We also ablate the confidence threshold $\tau\!\in\!\{0.2,0.3,0.4\}$; see Appendix~\ref{appendix:ablation} for details.

\paragraph{Soft aggregation for the marginal output distribution.}
A final adaptation addresses the estimation of the marginal output distribution $P(Y)$.
Discretizing predictions via argmax and then computing frequencies can introduce a measurement artifact:
for a totally uncertain (uniform) output, argmax resolves ties in a fixed manner (e.g., always selecting the first label),
artificially converting uncertainty into a sharp peak upon aggregation.
To avoid this, we use soft aggregation and estimate $P(Y)$ by averaging the regularized predictive distributions
across all $N$ samples:
\begin{align}
P(Y)=\frac{1}{N}\sum_{i=1}^{N}\hat{P}_i(Y),
\end{align}
where $\hat{P}_i(Y)\!=\!\hat{P}(Y \mid \cdot)_i$ denotes the regularized categorical distribution for sample $i$.
This preserves the model's output statistics and leads to a more faithful PID analysis.

\subsection{Analysis dimensions \& experimental settings}
\label{sec:exp_design}

To conduct a comprehensive information-decomposition analysis, 
we design experiments across three dimensions: (1) a large-scale comparison across a 
wide range of models and tasks, (2) the layer-wise information flow inside representative models,
and (3) the learning dynamics by examining model checkpoints 
throughout the training process.
For reproducibility, all experimental settings, including 
inference details and key hyperparameters, are provided in Appendix~\ref{appendix:setting}.

\subsubsection{Cross-model and cross-task comparison}

\paragraph{Models.} To assess how architecture and scale affect information 
use, we analyze 26 models (0.5B to 90B parameters) from 11 open-source 
LVLM families\footnote{All model checkpoints are downloaded from Hugging Face: \url{https://huggingface.co/models}.}. Our selection prioritizes recent, state-of-the-art 
families including LLaVA-OneVision~\citep{li2024llavaov}, 
Qwen2-VL~\citep{wang2024qwen2}, Qwen2.5-VL~\citep{bai2025qwen25}, 
Gemma-3~\citep{team2025gemma3}, InternVL2.5~\citep{chen2024internvl25}, 
InternVL3~\citep{zhu2025internvl3}, and Llama-3.2-Vision~\citep{meta2024llama31}. 
We also include Cambrian-1~\citep{tong2024cambrian} for its multi-vision encoder 
design and established models like LLaVA-1.5~\citep{liu2024llava15}, 
InstructBLIP~\citep{dai2023instructblip}, and Fuyu~\citep{fuyu-8b} to serve as baselines.

\paragraph{Tasks.} We evaluate all models on four diverse 
MC-VQA datasets: MMBench (`en\_dev')~\citep{liu2024mmbench} for general reasoning, 
POPE (`COCO14 adversarial')~\citep{li2023evaluating} and 
Reefknot (`Perception \& MCQ')~\citep{zheng2024reefknot} for 
hallucination evaluation, and PMC-VQA (`test\_clean')~\citep{zhang2023pmc} 
for domain-specific medical knowledge. Table~\ref{tab:datasets} summarizes the characteristics.

\begin{table}[htbp]
\caption{Details of the datasets used for evaluation. 
The listed training and test splits are not for LVLM fine-tuning; 
they are created by randomly partitioning each dataset (3:1 ratio) for the 
PID estimation, as the BATCH estimator requires separate sets to train networks and estimate PID values.}
\label{tab:datasets}
\centering
\begin{tabular}{llccc}
\Xhline{0.8pt}
\multicolumn{1}{c}{\bf Dataset} & \multicolumn{1}{c}{\bf Task} & \multicolumn{1}{c}{\bf \makecell{\# of \\ options}} & \multicolumn{1}{c}{\bf \makecell{\# of training \\ samples}} & \multicolumn{1}{c}{\bf \makecell{\# of test \\ samples}} \\
\hline
MMBench & General visual reasoning & 2--4 & 3246 & 1083 \\
POPE & Hallucination evaluation & 2 & 2250 & 750 \\
Reefknot & Hallucination evaluation & 4 & 1612 & 538 \\
PMC-VQA & Domain-specific knowledge & 4 & 1500 & 500 \\
\Xhline{0.8pt}
\end{tabular}
\end{table}


\paragraph{Image-removal intervention.}
As a behavioral validation, we remove the image to obtain a text-only baseline and measure the 
accuracy drop $D_{\text{vision}}$, which we relate to the PID-based information spectrum to assess models' visual reliance.

\subsubsection{Layer-wise information dynamics}
To trace the internal information flow, we conduct a layer-wise PID 
analysis on three representative model 
families: InternVL3-2B/8B, Qwen2.5-VL-3B/7B, and LLaVA-1.5-7B/13B. 
We analyze them on MMBench and PMC-VQA to observe dynamics on general 
and domain-specific tasks. 
By applying the \textit{logit lens}~\citep{nostalgebraist2020logitlens}, 
we project the hidden state at each transformer block through 
the LM head to obtain a layer-specific output distribution for our PID analysis.

\subsubsection{Learning dynamics of multimodal fusion}

To understand how fusion capabilities evolve, we analyze the two-stage 
training process of a representative model, 
LLaVA-1.5 (7B/13B), and reproduce its training 
using the original data and official settings. 
This process involves (1) vision-language alignment pretraining, where only 
the projector is trained to align frozen vision and 
language embeddings, followed by (2) visual instruction tuning, 
which fine-tunes both the projector and the LLM.

We save four equidistant checkpoints from each stage and evaluate 
each checkpoint's full PID profile on MMBench and PMC-VQA to 
create a temporal trace of its learning trajectory.

\section{Results and findings}
\label{res}

We treat PID components as signals: $R$ (overlap), $U_1$ (vision-only 
cues), $U_2$ (language-side knowledge), and $S$ 
(combined use). Across \textbf{breadth (models~$\times$~datasets)}, \textbf{depth (layers)}, and \textbf{time (training)}, we ask which signal most consistently shapes LVLM behavior and generalization.

\subsection{Dimension 1: Cross-model \& cross-task comparison}

Because redundancy $R$ and vision uniqueness $U_1$ are consistently small, 
we focus on synergy $S$ and language uniqueness $U_2$ to characterize task demand and 
model strategy (see full spectra in Appendix~\ref{appendix:d1}).

\subsubsection{Two regimes of information use across tasks}

\paragraph{Task-level patterns: two regimes of evidence use.}
\textit{Question: Do datasets push LVLMs to rely more on combining inputs, 
or on what they already know from text?}
To understand how different tasks challenge LVLMs, we first 
investigate whether they systematically demand different kinds of 
information. Our analysis examines how models allocate information between $S$ and $U_2$ on 
each dataset, revealing recurring differences that we summarize as two regimes of information use.




\begin{figure}[H]
\begin{minipage}[H]{0.50\textwidth}
Figure~\ref{fig:regime} plots, for each dataset, the dataset-level mean shares of $S$ and $U_2$ 
averaged across all models, with 95\% bootstrap CIs. A clear split emerges, 
driven mainly by $S$: MMBench and POPE form one cluster characterized by high $S$, 
while Reefknot and PMC-VQA form a second cluster with markedly lower 
$S$ and higher $U_2$. These are empirical profiles of how LVLMs 
behave on these datasets, not labels of the datasets themselves. 
In this second cluster, synergy appears to have a practical ceiling, 
suggesting that while fusion is beneficial, it cannot fully 
compensate for missing language-side knowledge; 
correspondingly, accuracies are typically 20--30\% lower than in the high-$S$ group.
\end{minipage}\hfill
\begin{minipage}[H]{0.47\textwidth}
\centering
\includegraphics[width=0.67\linewidth]{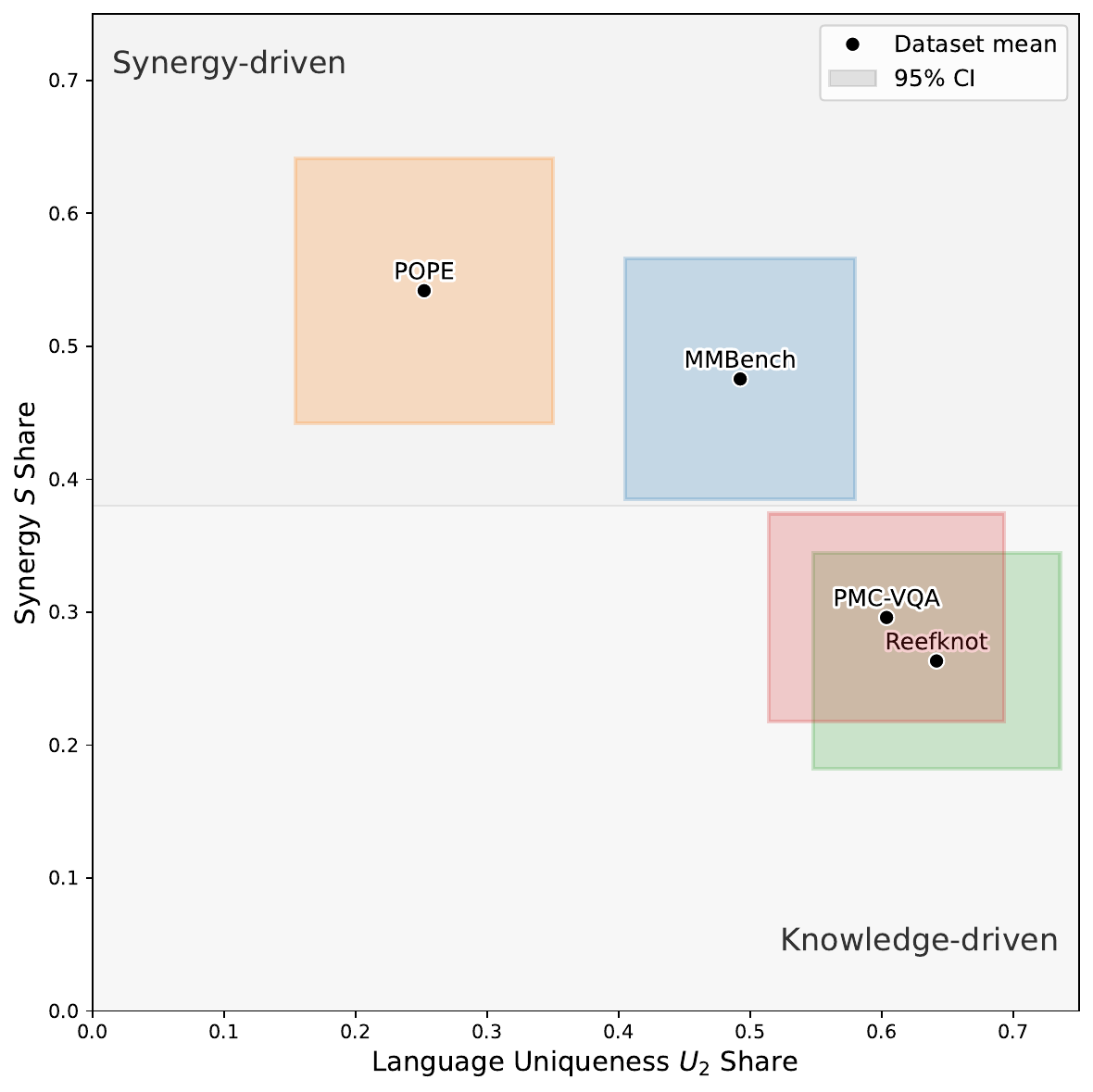}
\caption{Share of synergy $S$ and language uniqueness $U_2$ across four datasets.}
\label{fig:regime}
\end{minipage}
\end{figure}

\begin{finding}{1}
Across the four benchmarks, we observe two recurring information-use regimes in our experiments. 
A \textbf{synergy-driven regime} appears on tasks where models rely more on strong cross-modal fusion. 
In contrast, a \textbf{knowledge-driven regime} appears on tasks where performance reflects a stronger 
dependence on language-side knowledge and priors.
\end{finding}

\paragraph{Information correlates of accuracy across task regimes.}

\textit{Within each regime, we test which component tracks model accuracy.} 
We compute Spearman's $\rho$ between accuracy and each PID term across 26 LVLMs per dataset. 
Note that PID is computed from model predictions only; labels are used solely to report accuracy.


\begin{table}[htbp]
\caption{Spearman correlations ($\rho$) and $p$-values across datasets.}
\label{tab:spearman-simple}
\begin{center}
\begin{tabular}{lcccccccc}
\Xhline{0.8pt}
\multicolumn{1}{c}{\bf Dataset} &
\multicolumn{2}{c}{\bf $S$} &
\multicolumn{2}{c}{\bf $U_2$} &
\multicolumn{2}{c}{\bf $I(X_1,X_2;Y)$} &
\multicolumn{2}{c}{\bf $I(X_1;X_2;Y)$} \\
\multicolumn{1}{c}{} & {\bf $\rho$} & {\bf $p$-val} & {\bf $\rho$} & {\bf $p$-val} & {\bf $\rho$} & {\bf $p$-val} & {\bf $\rho$} & {\bf $p$-val} \\
\hline
MMBench & \textbf{0.750} & $<0.001$ & 0.194 & 0.343 & \textbf{0.632} & $<0.001$ & \textbf{-0.757} & $<0.001$ \\
POPE   & 0.742 & $<0.001$ & -0.009 & 0.964 & 0.157 & 0.445 & -0.701 & $<0.001$ \\
\hline
Reefknot & 0.357 & 0.073 & 0.313 & 0.119 & 0.266 & 0.196 & -0.348 & 0.081 \\
PMC-VQA & 0.432 & 0.027 & \textbf{0.406} & 0.040 & 0.559 & 0.003 & -0.587 & 0.002 \\
\Xhline{0.8pt}
\end{tabular}
\end{center}
\end{table}



Table~\ref{tab:spearman-simple} shows a consistent pattern. On synergy-driven benchmarks (MMBench, POPE),
$S$ is the strongest positive correlate of accuracy ($\rho\!\approx\!0.75$, $p\!<\!0.001$),
whereas $I(X_1,X_2;Y)$ is less consistent across datasets.\footnote{The interaction information
$I(X_1;X_2;Y)=R-S$ is strongly negative here, largely mirroring the $-S$ term.}
This implies that top-performing models are not those with simply ``more'' information, but those that
translate overlapping cues into effective cross-modal use.

On knowledge-driven benchmarks (Reefknot, PMC-VQA), the picture shifts:
$U_2$ becomes comparatively more informative (significant on PMC-VQA),
while $S$ remains positively related to accuracy but is no longer dominant.
These results suggest that fusion is beneficial in both regimes, but gains are bounded when language-side
knowledge becomes the primary bottleneck.


\begin{finding}{2}
In synergy-driven tasks, accuracy is most strongly associated with synergy $S$.
In knowledge-driven tasks, language uniqueness $U_2$ becomes more predictive, 
while synergy $S$ remains helpful but less dominant.
\end{finding}



\paragraph{Intervention-based validation: image removal.}
We further validate this interpretation via a simple intervention: removing the image and measuring the accuracy drop $D_{\text{vision}}$.
Across models, $D_{\text{vision}}$ correlates strongly with $S$ on synergy-driven benchmarks
(MMBench/POPE: $\rho=0.809/0.744$, both with $p<0.001$),
and more weakly on knowledge-driven ones (Reefknot/PMC-VQA: $\rho=0.459/0.400$, $p=0.018/0.043$).
This confirms a key prediction: models with higher $S$ are more sensitive to visual ablation, 
indicating that $S$ captures decision-relevant visual reliance.

Qualitative examples illustrating $S$-dominant (MMBench/POPE) and $U_2$-bounded (Reefknot/PMC-VQA) 
cases are provided in Appendix~\ref{appendix:case}.
We next ask whether such accuracy-relevant components reflect consistent model-level strategies.

\subsubsection{Information strategies across model architectures}

\begin{figure}[h]
    \centering
    \includegraphics[width=0.75\textwidth]{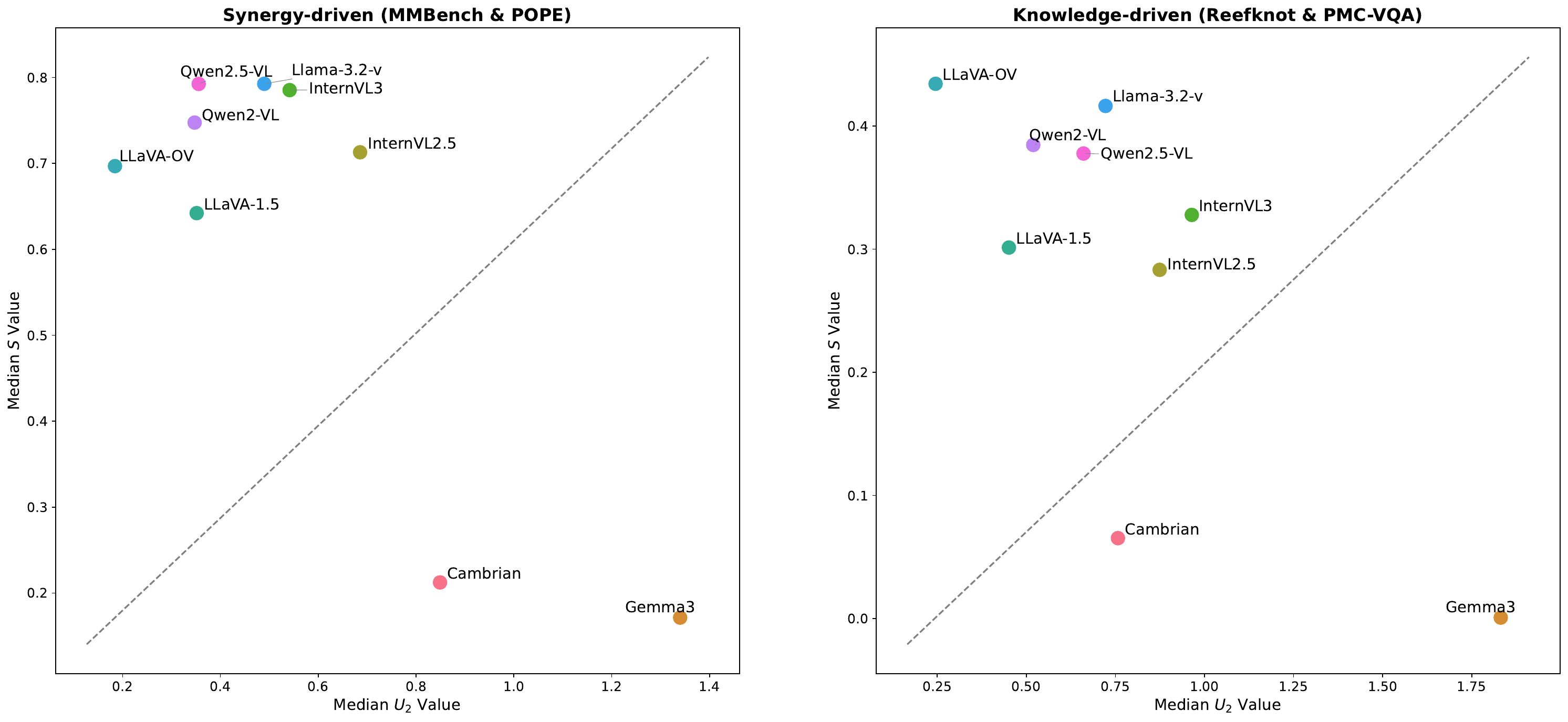}
    \caption{Family-level strategies: median $S$ versus median $U_2$ per family, computed across model sizes within each task regime. Points show the family medians for each regime. Outliers (InstructBLIP, Fuyu) are omitted for clarity.}
    \label{fig:family-strategy}
\end{figure}

\paragraph{Model families exhibit stable, contrasting information strategies.}
\textit{Do model families lean toward combining inputs or toward language knowledge---and does that preference hold across settings?}
We summarize each family's behavior by its median $S$ and $U_2$ within each regime, 
shown in Figure~\ref{fig:family-strategy}.

Families occupy two clearly separated regions, 
corresponding to two information-use strategies: a \textbf{fusion-centric} group 
(e.g., InternVL2.5/3, Qwen2/2.5-VL) with relatively high $S$ and lower $U_2$, 
and a \textbf{language-centric} group (e.g., Gemma3, Cambrian) with lower $S$ and higher $U_2$. 
Although absolute $S$ drops on knowledge-driven tasks, the relative positions of families remain 
similar across regimes, suggesting that this preference is a stable family-level tendency.


\begin{finding}{3}
LVLM families exhibit two stable information strategies:
\textbf{fusion-centric} families rely more on synergy $S$, while \textbf{language-centric} families rely more on language uniqueness $U_2$.
This strategic identity persists across task regimes.
\end{finding}

\paragraph{Scaling effects on synergy-driven tasks.}

\textit{If family identity is stable, scaling should reinforce the same leaning. A common expectation is that larger models rely more on $U_2$; we test this on synergy-driven tasks.} 
To assess how information use changes with scale, we compare Small (S), Mid (M), and Very-Large (VL) models within representative families.

\begin{table}[t]
\caption{Scaling on synergy-driven tasks: changes in accuracy ($\Delta$Acc) and PID shares ($\Delta S$, $\Delta U_2$) for S$\to$M and M$\to$VL within representative families.}
\label{tab:scaling-deltas}
\begin{center}
\begin{tabular}{l l r r r r r r}
\Xhline{0.8pt}
\multicolumn{1}{c}{\bf Family} &
\multicolumn{1}{c}{\bf Sizes (B)} &
\multicolumn{3}{c}{\bf S$\to$M (\%)} &
\multicolumn{3}{c}{\bf M$\to$VL (\%)} \\
 & & {\bf $\Delta$Acc} & {\bf $\Delta S$} & {\bf $\Delta U_2$} &
     {\bf $\Delta$Acc} & {\bf $\Delta S$} & {\bf $\Delta U_2$} \\
\hline
LLaVA-OneVision & 0.5$\to$7$\to$72 & 11.9 & 11.9 & -6.5 & 3.1 & 14.8 & -9.7 \\
Qwen2-VL        & 2$\to$7$\to$72   &  5.7 &  0.6 &  8.4 & -3.9 & 0.4 & -0.6 \\
Qwen2.5-VL      & 3$\to$7$\to$72   &  1.3 & -6.3 &  1.0 & 2.5 & 5.5 &  2.5 \\
InternVL2.5     & 2$\to$8$\to$78   &  7.3 &  36.8 & -55.6 & 3.6 & 10.6 & 3.8 \\
InternVL3       & 2$\to$8$\to$78   &  2.7 &  2.5 & -6.2 & 6.4 & 4.6 & -10.3 \\
\Xhline{0.8pt}
\end{tabular}
\end{center}
\end{table}

As shown in Table~\ref{tab:scaling-deltas}, the share of language uniqueness $U_2$ does not 
systematically increase with size and often decreases. By contrast, the accuracy differences 
between sizes tend to co-vary with changes in the share of synergy $S$: larger checkpoints that 
improve more in accuracy also exhibit larger increases in $S$. This is consistent with 
Finding~2, where on synergy-driven tasks performance is more closely tied to $S$ than to $U_2$.

\begin{finding}{4}
On synergy-driven tasks, within each family the accuracy differences between available sizes 
are more closely associated with changes in $S$ than with changes in $U_2$. 
Larger checkpoints therefore tend to benefit more from stronger multimodal 
fusion than from further amplifying language-side priors.
\end{finding}

\subsection{Dimension 2: Layer-wise information dynamics}

\textit{Because stable family preferences and gains with scale in Dimension 1 tracked increases in $S$, the next question is where $S$ arises in the stack.}
We therefore analyze layer-wise information with the \textit{logit lens}; full results are in Appendix~\ref{appendix:d2}.

\begin{figure}[h]
    \centering
    \includegraphics[width=0.95\textwidth]{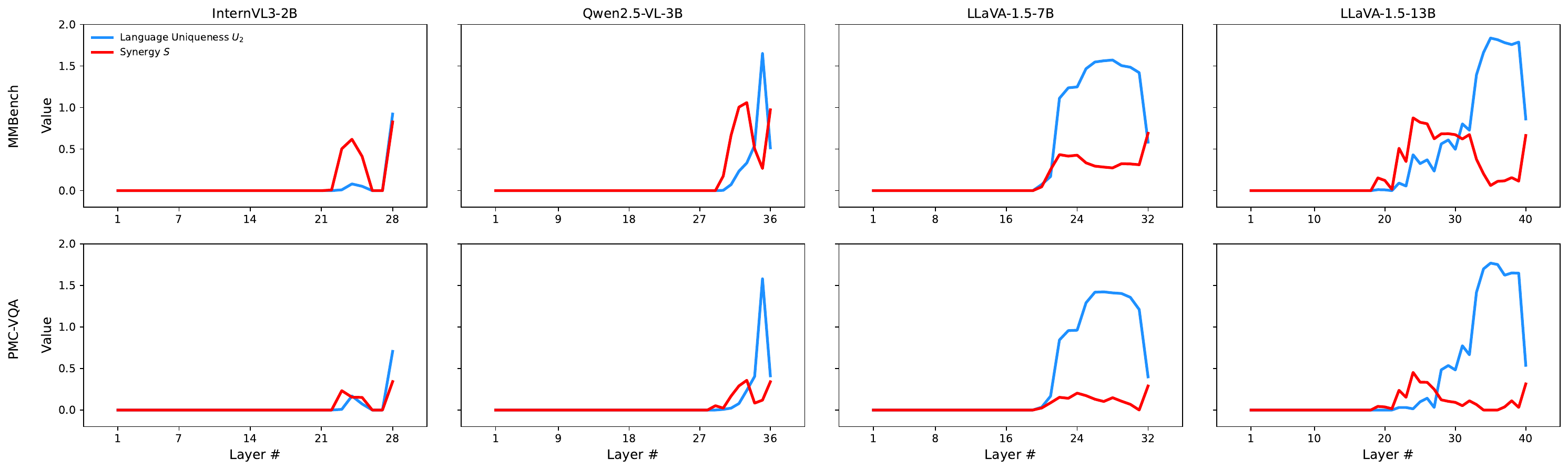}
    \caption{Layer-wise PID dynamics for representative models on synergy-driven (MMBench, top) and knowledge-driven (PMC-VQA, bottom) tasks. A consistent three-phase pattern appears across models and datasets.}
    \label{fig:layer-wise}
\end{figure}

For Qwen2.5-VL-7B and InternVL3-8B, since the logit lens does not yield meaningful intermediate predictions, we omit them here.
Figure~\ref{fig:layer-wise} shows a consistent three-phase profile across models and datasets. 
$S$ typically emerges and peaks in the middle-to-late layers, often softens near the output, then spikes at the final layer. 
$U_2$ generally builds through the stack, peaking at the second-to-last layer before a sharp final drop. 
$R$ and $U_1$ remain small throughout. 
An exception is InternVL3-2B, where $U_2$ does not exhibit the final drop.

\begin{finding}{5}
The layer-wise dynamics reveal a shared three-phase pattern of information flow. 
Information emerges in the middle-to-late layers, then moves from language-based 
representation building in the later layers to a decisive, synergistic fusion 
of modalities in the final layer.
\end{finding}

\subsection{Dimension 3: Learning Dynamics of Multimodal Fusion}

\textit{While layer-wise snapshots reveal \textbf{where} $S$ arises, they do not show \textbf{when} it emerges. We therefore turn to the training trajectory.}
We trace PID through the two-stage training of LLaVA-1.5 (7B, 13B). Full results are in Appendix~\ref{appendix:d3}.

\begin{figure}[H]
\begin{minipage}[H]{0.50\textwidth}
The results in Figure~\ref{fig:learning-dynamics} show a clear separation between the two 
training stages in our reproduced LLaVA-1.5 pipeline. Throughout alignment pretraining 
(Stage~1), both $S$ and $U_2$ remain low and relatively stable. Once visual instruction 
tuning begins (Stage~2), both components increase markedly. The effect of model scale 
also differs across components: the 7B model shows a more pronounced increase in $S$, 
whereas the 13B model exhibits a stronger increase in $U_2$, indicating that larger models 
in this setting place greater emphasis on language-side priors during fine-tuning. Overall, 
both trends suggest that fine-tuning benefits from scale.
\end{minipage}\hfill
\begin{minipage}[H]{0.45\textwidth}
\centering
\includegraphics[width=0.98\linewidth]{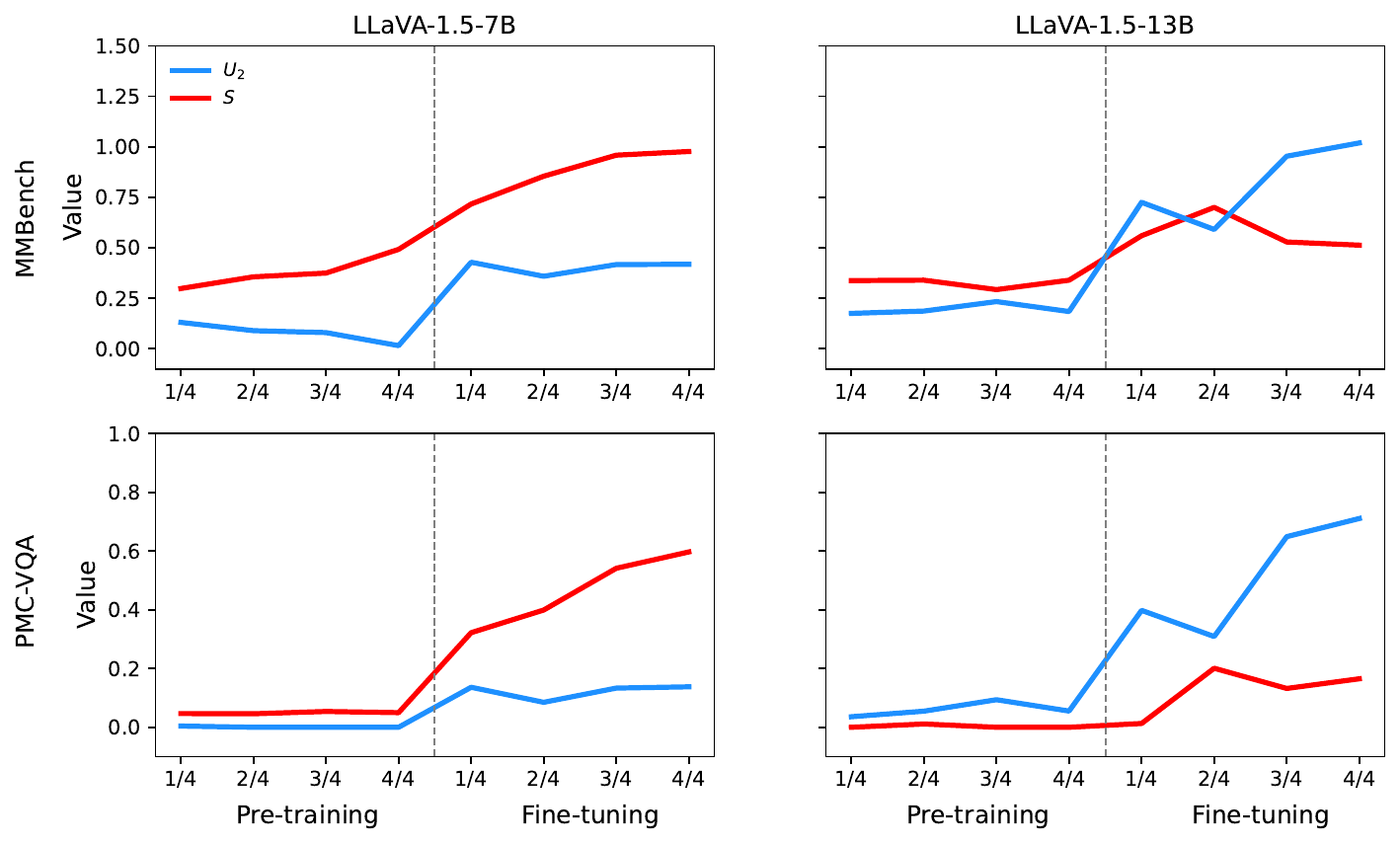}
\caption{Evolution of $S$ and $U_2$ during two-stage training of LLaVA-1.5 (7B, 13B).}
\label{fig:learning-dynamics}
\end{minipage}
\end{figure}

This aligns with observations from prior work such as MiniGPT-4~\citep{zhu2023minigpt}, 
where the second-stage visual instruction tuning was found to be crucial for improving 
generation quality. Our PID analysis provides an information-theoretic view of this 
phenomenon, showing how fusion $S$ and language priors $U_2$ emerge and diverge across 
stages and model scales.

\begin{finding}{6}
In the two-stage training of LLaVA-1.5, multimodal fusion $S$ remains negligible during 
alignment pretraining and increases primarily during visual instruction tuning. 
Across the two available sizes, the 7B model shows relatively larger gains in $S$, 
whereas the 13B model exhibits comparatively stronger growth in $U_2$ during this stage.
\end{finding}


\paragraph{Summary of empirical findings.} Taken together, our results show that PID provides a coherent view of LVLM behavior across 
three complementary axes. At the task level, benchmarks fall into recurring information-use regimes, 
and at the family level, model families adopt contrasting strategies characterized by different balances of $S$ and $U_2$; 
at the layer level, $S$ and $U_2$ follow a shared three-phase pattern of information flow; and across training stages, 
multimodal fusion $S$ emerges primarily during visual instruction tuning and interacts with model scale.

\section{Conclusion}
\label{sec:conclusion}

LVLMs are typically evaluated by accuracy, which tells us \emph{what} they get right but not 
\emph{how} different modalities are utilized. In this work, we introduce a PID-based framework that yields a 
process-level decomposition of decision-relevant information in LVLMs and, to our knowledge, provides the first systematic application of PID at this scale. 
By adapting a scalable PID estimator to LVLM outputs and applying it to 26 models across four benchmarks, we offer an information-theoretic lens that complements accuracy-only evaluation and supports more targeted analysis of multimodal behavior.

This study has several \textbf{limitations}. (1) PID estimation assumes a discrete target space, so we do not cover fully open-ended generation tasks. 
(2) Our unimodal probes are approximate: masking a modality with calibrated noise stabilizes estimation, but $U_1$, $U_2$, and $S$ are measured under this probe rather than under truly natural unimodal inputs. 
(3) PID is correlational: the components are derived from model predictions and inputs, and their relationships to accuracy or interventions reflect associations rather than full causal mechanisms.


\textbf{Future work} can extend this study in several directions:
\begin{enumerate}[label=\arabic*., leftmargin=*, itemsep=0.2em, topsep=0.2em]
    \item \textbf{Methodology:} developing PID estimators and output encodings that handle richer generative settings and additional modalities, and exploring complementary unimodal probes.
    \item \textbf{Model and training design:} using $(U_1, U_2, S)$ as diagnostic signals during scaling and instruction tuning, and potentially as auxiliary objectives to balance fusion and language priors.
    \item \textbf{Evaluation:} using PID-based analyses to guide the construction of benchmarks that explicitly require high synergy $S$ or isolate language priors $U_2$.
\end{enumerate}

\newpage


\subsubsection*{Acknowledgments}
This work was supported by JST-SPRING Grant Number JPMJSP2108, JST-CRONOS Grant Number JPMJCS24K8, JSPS KAKENHI Grant 
Number JP23K28139, and the Institute of AI and Beyond of the University of Tokyo.

\subsubsection*{Ethics statement}
All authors have read and follow the ICLR Code of Ethics. Our study analyzes publicly 
available LVLM checkpoints and datasets (MMBench, POPE, Reefknot and PMC-VQA); no new 
data were collected, no human subjects were involved, and no personally identifiable 
information is used. We respect dataset/model licenses and cite original sources; 
we do not redistribute proprietary content. The work focuses on 
interpretability/analysis (PID of information use) rather than deployment, and is 
intended to improve transparency of multimodal systems. 
Potential risks (e.g., inherited dataset or model biases) are acknowledged; 
we report cases in the paper/appendix. 
The authors declare no conflicts of interest and no sponsorship that influenced the results.

\subsubsection*{Reproducibility statement}
Our proposed PID framework and experimental design are described in Section~\ref{method}.
Full tables/plots and additional analyses appear in Appendix~\ref{appendix:allresults} and 
robustness checks in Appendix~\ref{appendix:ablation}. 
The public repository linked in the abstract contains code and configuration 
files needed to rerun the study, including dataset splits, prompts, preprocessing steps, 
model versions, and random seeds, as well as scripts to compute PID.

\bibliography{references}
\bibliographystyle{iclr2026_conference}

\newpage

\appendix

\section*{LLM usage}
We used an LLM-based assistant only for writing support, including sentence paraphrasing, 
grammar/typo checking, and outline/flow refinement. 
The LLM was not used to design experiments, analyze data, or 
generate results; all technical content (methods, equations, figures, and code) was 
produced and verified by the authors. LLM-suggested text was reviewed and 
edited for accuracy, and references were inserted and checked manually.

\section{Partial information decomposition and its estimation}
\label{appendix:pid}

\subsection{Introduction of partial information decomposition}
\label{appendix:pid:intro}

Classical mutual information (MI) quantifies the information a single 
variable provides about another: $I(X;Y)$ measures the reduction in 
uncertainty from $H(Y)$ to $H(Y \mid X)$ \citep{shannon1948mathematical}. 
Extending this analysis to a system with multiple 
source variables---for instance, two sources $X_1$ and $X_2$ with state 
spaces $\mathcal{X}_1$ and $\mathcal{X}_2$, and a target $Y$ with 
state space $\mathcal{Y}$---is challenging, as the standard 
interaction information, $I(X_1;X_2;Y)$, can be positive or negative. 
This sign ambiguity complicates its interpretation and 
motivates a decomposition of information into a set of well-behaved, non-negative quantities.

Partial information decomposition (PID), first proposed by \citet{williams2010nonnegative}, 
addresses this issue. We adopt the definition from \citet{bertschinger2014quantifying}, 
which decomposes the total 
information into three conceptual components: redundancy, uniqueness, and synergy. 
These concepts are quantified by four non-negative atoms: redundant information ($R$), 
unique information from the first source ($U_1$), unique information 
from the second source ($U_2$), and synergistic information ($S$).

PID postulates the following consistency relations linking the four 
atoms to 4 classical mutual information terms:
\begin{align}
I(X_1,X_2;Y) &= R + U_1 + U_2 + S, \label{eq:pid-sum}\\
I(X_1;Y) &= R + U_1, \label{eq:pid-x1}\\
I(X_2;Y) &= R + U_2, \label{eq:pid-x2}\\
I(X_1;X_2;Y) &:= I(X_1;Y)+I(X_2;Y)-I(X_1,X_2;Y) = R - S. \label{eq:pid-coinfo}
\end{align}
Eqs.~\ref{eq:pid-sum}--\ref{eq:pid-coinfo} establish the algebraic relationship between $(R,U_1,U_2,S)$ and 
the usual information measures, 
cleanly separating redundancy ($R$) from synergy ($S$) through 
the co-information identity Eq.~\ref{eq:pid-coinfo}.

The definition for the PID atoms relies on optimization over the set of \emph{marginal-matching} distributions:
\begin{align}
\Delta_{P} = \bigl\{\, Q \in \Delta : Q(x_i,y) = P(x_i,y),\ \forall\, x_i \in \mathcal{X}_i,\ y \in \mathcal{Y},\ i \in \{1,2\} \,\bigr\},
\end{align}
where $\Delta$ is the space of all possible joint distributions of $X_1$, $X_2$ and $Y$. 
This set contains all distributions $Q$ that preserve the 
marginal information of the true distribution $P$, while allowing 
other dependencies to vary. 
Let $I_Q(\cdot)$ denote mutual information under a distribution $Q$, 
the PID atoms are defined as:
\begin{align}
R &= \max_{Q \in \Delta_P} I_Q(X_1; X_2; Y), \label{eq:R-opt}\\
U_1 &= \min_{Q \in \Delta_P} I_Q(X_1; Y  \mid  X_2), \label{eq:U1-opt}\\
U_2 &= \min_{Q \in \Delta_P} I_Q(X_2; Y  \mid  X_1), \label{eq:U2-opt}\\
S &= I(X_1, X_2; Y) - \min_{Q \in \Delta_P} I_Q(X_1, X_2; Y). \label{eq:S-opt}
\end{align}
Intuitively, optimizing over $\Delta_P$ isolates each 
component of information. For instance, 
a distribution $Q^\star \in \Delta_P$ that minimizes the 
total mutual information $I_Q(X_1,X_2;Y)$ does so by reducing the higher-order (synergistic/complementary) dependencies while preserving the source--target marginals. 
The resulting gap, used to define synergy in Eq.~\ref{eq:S-opt}, 
therefore measures this emergent information. 
Similarly, the unique information components (Eqs.~\ref{eq:U1-opt} and \ref{eq:U2-opt}) represent 
the minimal necessary conditional information from 
each source, while redundancy (Eq.~\ref{eq:R-opt}) represents the maximal possible shared co-information.

Under this construction, the four atoms are non-negative and satisfy 
the axiomatic relations in Eqs.~\ref{eq:pid-sum}--\ref{eq:pid-coinfo}. 
A key insight from \citet{bertschinger2014quantifying} is that the 
optimization problems defining $R$, $U_1$, $U_2$, and $S$ are \emph{equivalent}, and it is sufficient to solve one of them.

\subsection{The estimator we leverage: BATCH for continuous representations}
\label{appendix:pid:batch}

We adopt BATCH, a scalable PID estimator for high-dimensional, continuous input representations $X_1$ and $X_2$ and 
large datasets \citep{liang2023}. The method amortizes the optimization problems in Eqs.~\ref{eq:R-opt}--\ref{eq:S-opt} by 
learning a parametric joint $\tilde{Q}(x_1,x_2,y)$ that 
lies in the marginal-matching set $\Delta_{P}$ defined earlier. Specifically, $\tilde{Q}$ is trained to approximately solve
\begin{align}
    \min_{Q\in\Delta_{P}} I_{Q}(X_1,X_2;Y),
\end{align}
and, by the equivalence of the optimization characterizations, 
the remaining PID components are obtained by evaluating 
the corresponding quantities under the same $\tilde{Q}$.

\paragraph{Neural parameterization and projection.}
Given mini-batches $(\mathbf{X}_1,\mathbf{X}_2,\mathbf{Y})$, two encoders $f_{\phi(1)}$ and $f_{\phi(2)}$ output 
features that define an \emph{unnormalized} joint via a similarity matrix:
\begin{align}
A \;=\; \exp\!\big(f_{\phi(1)}(\mathbf{X}_1,y)\, f_{\phi(2)}(\mathbf{X}_2,y)^\top\big),
\end{align}
where $A[i][j][y] = \tilde{Q}(\mathbf{X}_1[i],\mathbf{X}_2[j],y)$ for each $y \in \mathcal{Y}$. 
To enforce the marginal 
constraints $Q\!\in\!\Delta_{P}$, BATCH applies the Sinkhorn-Knopp algorithm \citep{cuturi2013sinkhorn} to iteratively normalize rows and columns of $A$ 
so the projected distribution matches the fixed pairwise marginals $P(x_1,y)$ and $P(x_2,y)$.

\paragraph{Training objective.}
Given matrix $A$ representing $\tilde{Q}(x_1,x_2,y)$, the objective can be written as:
\begin{equation}
\min_{Q\in\Delta_P}\;
\mathbb{E}_{\substack{x_1,y\sim Q(x_1,y)\\ x_2\sim Q(x_2  \mid  x_1,y)}}
\left[
\log
\frac{ Q(x_2  \mid  x_1,y)\, Q(x_1  \mid  y) }
{ \sum_{y'\in Y} Q(x_2  \mid  x_1,y')\, Q(y'  \mid  x_1)\, Q(x_1) }
\right],
\label{eq:batch-loss-cap}
\end{equation}
hence gradient descent is leveraged to optimize this via 
updating the parameters of $f_{\phi(1)}$ and $f_{\phi(2)}$.

\paragraph{Estimating PID values via learned models.}
Upon convergence, we estimate the required information terms under the data distribution $P$ and the learned $\tilde{Q}$. 
Using the consistency relations and the optimization equivalence, 
the PID components are obtained as
\begin{align}
R \;&=\; I_{\tilde{Q}}(Y;X_1,X_2), \\
U_1 \;&=\; I_{\tilde{Q}}(Y;X_1,X_2) - I_{P}(Y;X_2), \\
U_2 \;&=\; I_{\tilde{Q}}(Y;X_1,X_2) - I_{P}(Y;X_1), \\
S \;&=\; I_{P}(Y;X_1,X_2) - I_{\tilde{Q}}(Y;X_1,X_2).
\end{align}
These quantities satisfy the PID consistency equations by 
construction and recover the optimization-defined 
components when $\tilde{Q}$ is learned.

\paragraph{Rationale for adopting the BATCH estimator.}
The BATCH estimator provides a practical and scalable method 
for calculating these PID atoms from data. The core of this approach is to use neural 
networks to parameterize and learn an approximate joint distribution, $\tilde{Q}$, that satisfies 
the required marginal-matching constraints. By optimizing an information-theoretic objective 
over mini-batches, the estimator can be effectively applied to large datasets. 
We chose to adapt this estimator for two primary reasons. First, it was explicitly 
designed for general multimodal learning contexts. Second, and most importantly, 
it operates on high-dimensional, continuous features. This latter property 
makes it uniquely suited for analyzing modern LVLMs, 
as our framework can apply it directly to the rich vector embeddings
these models produce to quantify their internal information dynamics.

\newpage

\section{Ablation study}
\label{appendix:ablation}

To validate our methodology, we examine sensitivity to two implementation choices: 
(i) feature summarization (mean pooling, last-hidden state, and max pooling) and (ii) the 
confidence threshold $\tau\!\in\!\{0.2,0.3,0.4\}$. 
We evaluate four representative LVLMs chosen to span \emph{families}, \emph{scales}, and \emph{strategy types}: 
Qwen2.5-VL-7B and Qwen2.5-VL-72B (fusion-centric, two scales of the same family), 
Gemma3-4B (language-centric), and Cambrian-34B (language-centric with more parameters). 
Ablations are run on the synergy-driven MMBench and the knowledge-driven PMC-VQA datasets. 
Because $S$ and $U_2$ are the primary components in these regimes, we report two summary 
tables: synergy $S$ on MMBench (Table~\ref{tab:ablation-s-mmb}) and 
language uniqueness $U_2$ on PMC-VQA (Table~\ref{tab:ablation-u2-pmc}).

\begin{table}[htbp!]
\caption{$S$ on MMBench for four chosen models under two ablations (feature summarization and confidence threshold).}
\label{tab:ablation-s-mmb}
\begin{center}
\begin{tabular}{lcccccc}
\Xhline{0.8pt}
\multicolumn{1}{c}{\bf Model} &
\multicolumn{3}{c}{\bf Feature summarization} &
\multicolumn{3}{c}{\bf Confidence threshold $\tau$} \\
 & {\bf Mean (ours)} & {\bf Last-hidden} & {\bf Max-pool} &
   {\bf 0.3 (ours)} & {\bf 0.2} & {\bf 0.4} \\
\hline
Qwen2.5-VL-7B  & 1.112 & 1.112 & 1.112 & 1.112 & 1.112 & 1.112 \\
Qwen2.5-VL-72B  & 1.088 & 1.088 & 1.088 & 1.088 & 1.088 & 1.088 \\
Gemma3-4B      & 0.167 & 0.172 & 0.173 & 0.167 & 0.167 & 0.167 \\
Cambrian-34B   & 0.630 & 0.637 & 0.630 & 0.630 & 0.606 & 0.630 \\
\Xhline{0.8pt}
\end{tabular}
\end{center}
\end{table}

\begin{table}[htbp]
\caption{$U_2$ on PMC-VQA for four chosen models under two ablations (feature summarization and confidence threshold).}
\label{tab:ablation-u2-pmc}
\begin{center}
\begin{tabular}{lcccccc}
\Xhline{0.8pt}
\multicolumn{1}{c}{\bf Model} &
\multicolumn{3}{c}{\bf Feature summarization} &
\multicolumn{3}{c}{\bf Confidence threshold $\tau$} \\
 & {\bf Mean (ours)} & {\bf Last-hidden} & {\bf Max-pool} &
   {\bf 0.3 (ours)} & {\bf 0.2} & {\bf 0.4} \\
\hline
Qwen2.5-VL-7B  & 0.665 & 0.665 & 0.665 & 0.665 & 0.665 & 0.665 \\
Qwen2.5-VL-72B  & 0.893 & 0.893 & 0.893 & 0.893 & 0.893 & 0.893 \\
Gemma3-4B      & 1.864 & 1.864 & 1.864 & 1.864 & 1.864 & 1.864 \\
Cambrian-34B   & 0.698 & 0.698 & 0.698 & 0.698 & 0.698 & 0.698 \\
\Xhline{0.8pt}
\end{tabular}
\end{center}
\end{table}

\paragraph{Input feature summarization.}
We compare mean pooling (used in the main experiments) with two common 
alternatives: \textbf{last-hidden state} and \textbf{max pooling}. 
On MMBench (Table~\ref{tab:ablation-s-mmb}), $S$ is unchanged for the two Qwen2.5-VL models; 
Gemma3-4B varies slightly ($0.167\!\to\!0.173$), and Cambrian-34B varies within $0.630$--$0.637$. 
On PMC-VQA (Table~\ref{tab:ablation-u2-pmc}), $U_2$ is identical across all pooling choices for all models. 
Thus, feature summarization has negligible effect on the components most relevant to each regime.

\paragraph{Confidence threshold $\tau$.}
We vary the regularization threshold around our default ($\tau\!=\!0.3$) to $\tau\!\in\!\{0.2,0.4\}$. 
On MMBench (Table~\ref{tab:ablation-s-mmb}), $S$ is invariant for both Qwen2.5-VL models and for Gemma3-4B; 
Cambrian-34B shows a small dip at $\tau\!=\!0.2$ ($0.606$ vs.\ $0.630$ at $\tau\!=\!0.3/0.4$). 
On PMC-VQA (Table~\ref{tab:ablation-u2-pmc}), $U_2$ is unchanged across $\tau$ for all models. 
These results indicate stability of our conclusions with respect to the confidence-regularization setting in the tested range.

\textbf{Summary.} Across both ablations, the regime-defining components ($S$ on MMBench, $U_2$ on PMC-VQA) remain 
effectively constant, supporting the robustness of our methodology.

\newpage

\section{Detailed experimental settings}
\label{appendix:setting}

All experiments reported in this paper, including model inference, 
PID estimation, and training reproduction, were conducted on servers equipped with 8 NVIDIA A100 GPUs.

\paragraph{General inference details.}
For all multiple-choice VQA tasks, we use a standardized prompt that instructs 
the model to answer with only the letter of 
the correct option: ``Please select the correct answer from the options above. 
You must answer with the letter of the correct option only.'' 
While most modern LVLMs adhere to this instruction, 
some earlier models (e.g., InstructBLIP, Fuyu-8b) tend to generate conversational, 
free-form text.

To handle these inconsistencies, our reported ``accuracy'' is not a 
standard logit-based metric. Instead, we perform a strict 
string match on the first generated token: a 
prediction is correct only if the normalized 
token (lowercased, punctuation removed) exactly matches the 
ground-truth letter. This format-dependent evaluation means 
the performance floor is 0\%, not the random-guess rate, as models that fail to follow the required format will be marked incorrect.
For reproducible outputs, we use a deterministic greedy decoding 
strategy (no sampling) for all models.

\paragraph{Hyperparameters for BATCH estimator.}
Although the BATCH estimator is known to be robust to hyperparameter 
settings \citep{liang2023}, we adhere to the original configuration for 
consistency and reproducibility. The key hyperparameters used for the 
estimator's neural networks are listed in Table~\ref{tab:batch_params}.
\begin{table}[h]
\caption{Hyperparameters for the BATCH Estimator.}
\label{tab:batch_params}
\centering
\begin{tabular}{ll}
\Xhline{0.8pt}
\textbf{Hyperparameter}      & \textbf{Value} \\
\hline
Learning rate                & 1e-3 \\
Optimizer                    & Adam \\
Number of epochs                       & 8 \\
Network architecture         & 3-layer MLP \\
Hidden dimension       & 32 \\
Activation function          & ReLU \\
Training batch size          & 256 \\
Test batch size              & 256 \\
\Xhline{0.8pt}
\end{tabular}
\end{table}

\paragraph{Hyperparameters for reproducing LLaVA-1.5 training.}
We exactly follow the official two-stage recipe of \citet{liu2024llava15} with
\textbf{no changes} to data, model, or optimization hyperparameters. For analysis, we saved 
four equally spaced checkpoints from each stage and evaluated
with greedy decoding (sampling disabled). For full hyperparameters, see Table~9 of \citet{liu2024llava15}.

\newpage

\section{Case studies}
\label{appendix:case}

To provide a more qualitative understanding of our findings, we 
visualize the PID results for two representative examples. 
For each VQA pair, we show the outputs from four LVLMs that 
exemplify different strategies (fusion-centric vs.\ language-centric) and scales. 
These cases provide concrete illustrations of how different 
models use information to solve tasks from the two distinct regimes we identified.

\paragraph{Case 1: A synergy-driven task (MMBench).}
This task requires the model to identify the state of Massachusetts 
on a map. 
Success depends on correctly associating the visual shape 
of the state with 
its name in the text---a 
classic fusion task where neither modality alone is sufficient.

\begin{figure}[h]
    \centering
    \includegraphics[width=0.95\columnwidth]{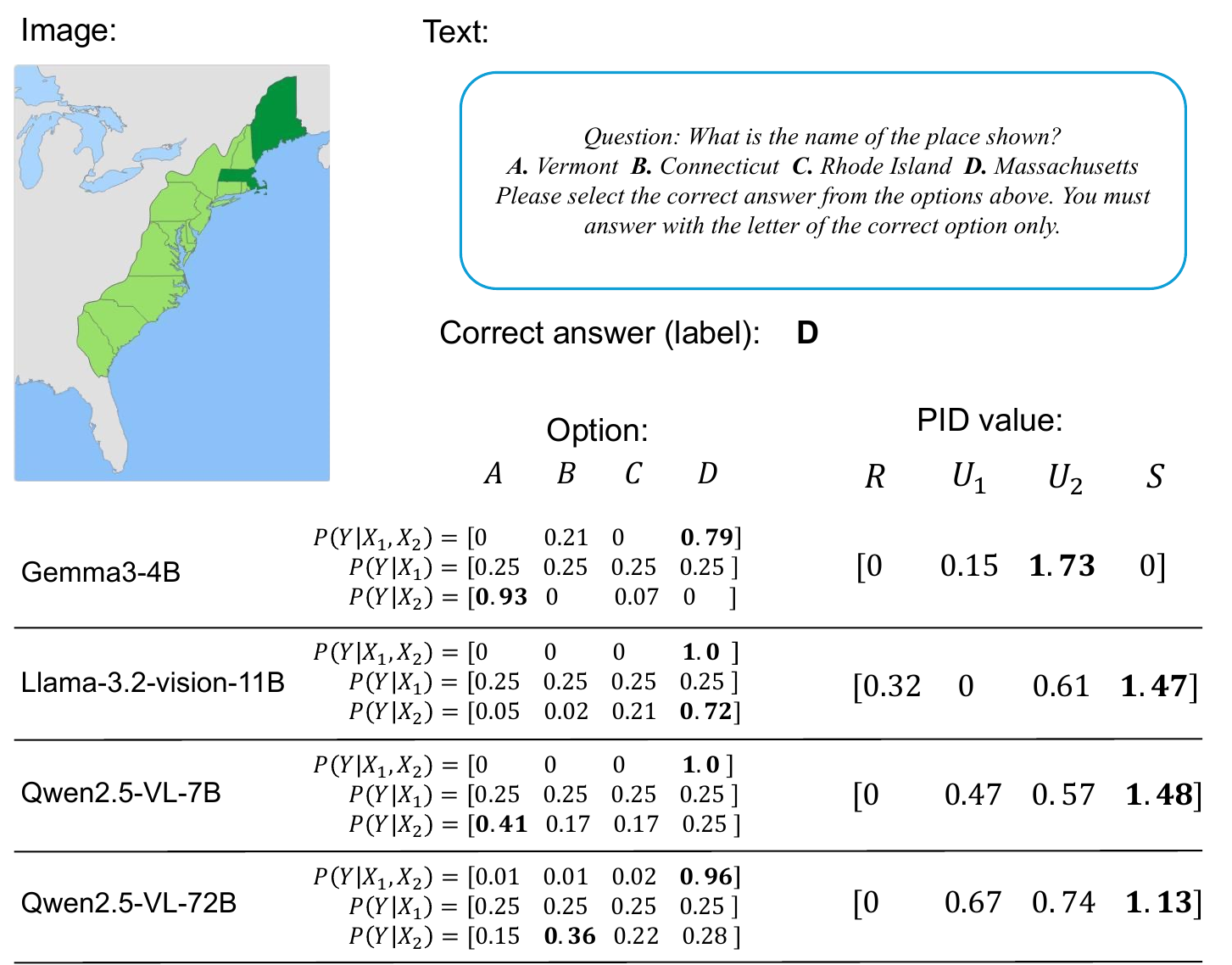}
    \caption{PID analysis for a synergy-driven task. All models answer correctly, 
    but the PID results reveal two distinct solution strategies: 
    generating high synergy (Llama-3.2-vision, Qwen2.5-VL) versus 
    correcting a strong, incorrect language prior (Gemma3).}
    \label{fig:case1}
\end{figure}

As shown in Figure~\ref{fig:case1}, all models arrive at the correct answer, 
but their methods differ dramatically. 
The fusion-centric models (Llama-3.2-vision and the Qwen2.5-VL series) solve 
the problem by generating a large amount of synergy $S$, confirming 
that the answer emerges from the direct interaction of image and text. 
In stark contrast, the language-centric Gemma model also answers correctly, 
but does so by overcoming a strong, incorrect language prior (a high $U_2$ favoring ``Vermont''). 
Interestingly, the PID framework reveals this correction happens without 
generating synergy, showcasing a different path to success: 
the visual information acts to override a mistaken language bias, 
rather than creating new information with it. 
This highlights the framework's ability to distinguish 
between models that truly \emph{fuse} modalities to create new insight and 
those that \emph{arbitrate} between conflicting unimodal beliefs. 
This latter ``correction'' mechanism, where visual evidence $U_1$ 
overrides a strong language bias $U_2$, may represent an efficient, 
non-synergistic strategy common in smaller or more language-centric architectures.

\clearpage

\paragraph{Case 2: A knowledge-driven task (Reefknot).}
This task asks about the spatial relationship ``through'', which 
requires a nuanced understanding of prepositions that primarily 
resides within the language model. The visual context is relatively 
simple, but the linguistic concept is complex.

\begin{figure}[h]
    \centering
    \includegraphics[width=0.95\columnwidth]{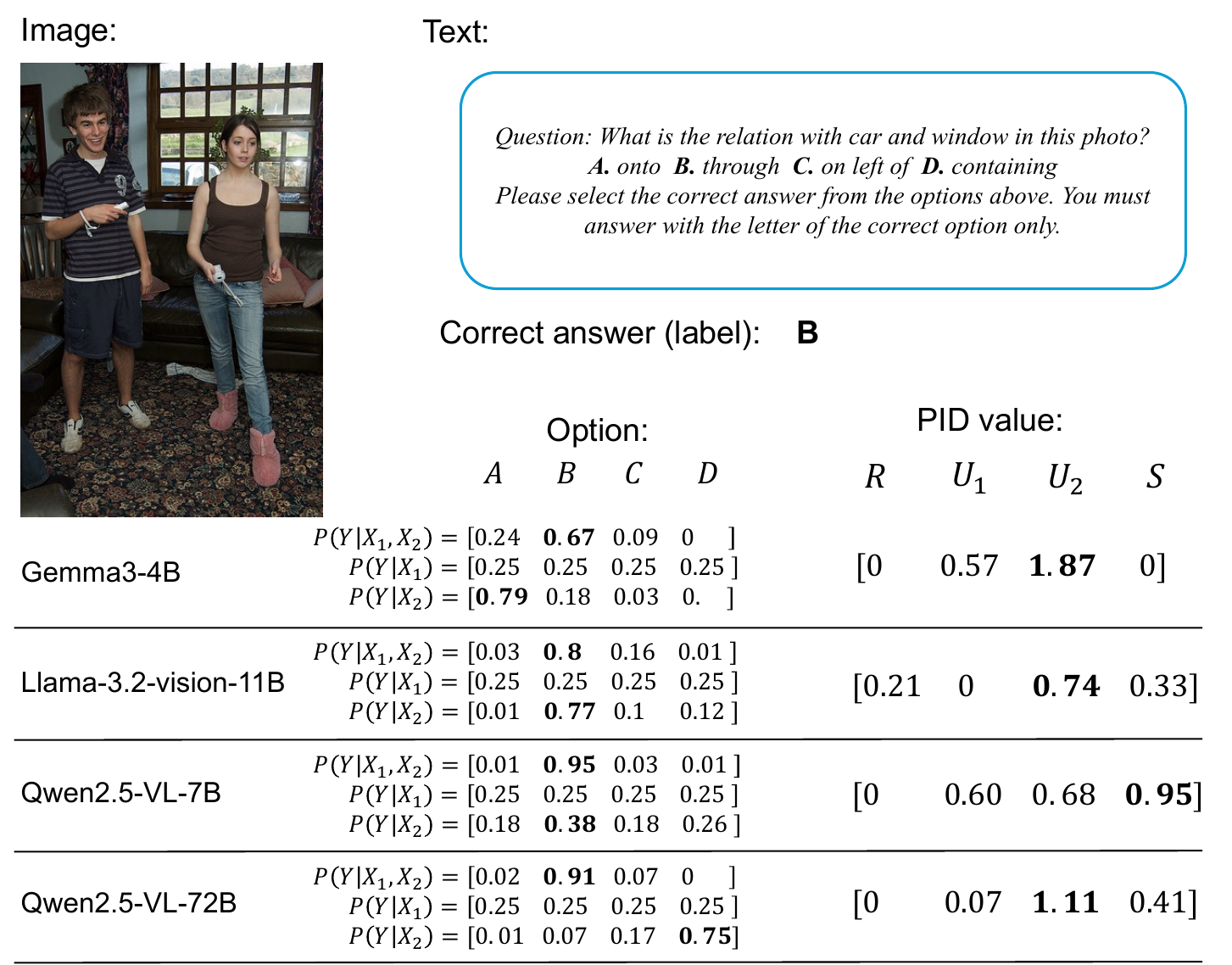}
    \caption{PID analysis for a knowledge-driven task. 
    While all models rely heavily on language uniqueness $U_2$, 
    fusion-centric models like Llama-3.2-vision and Qwen2.5-VL also 
    generate non-trivial synergy, unlike the language-centric Gemma.}
    \label{fig:case2}
\end{figure}

Figure~\ref{fig:case2} illustrates how these different strategies 
adapt to a knowledge-driven task, presenting a clear contrast to Case 1.
As expected, the language-centric Gemma model solves the problem almost 
exclusively with its language prior (high $U_2$), generating no synergy. 
The fusion-centric models (Llama-3.2-vision and the Qwen2.5-VL family), 
however, tell a more interesting story: while they also depend 
heavily on language uniqueness, they continue to 
generate significant synergy (e.g., $S=0.95$ for Qwen2.5-VL-7B).

This demonstrates a persistent strategic difference between model families. 
Even when a task seems solvable with language alone, 
fusion-centric models consistently attempt to integrate visual information 
to ground their linguistic understanding, 
whereas language-centric models default to their strong language priors.

These case studies provide qualitative evidence for our quantitative findings, 
visually demonstrating how models with different core strategies 
can reach the same correct answer via entirely 
different information-processing pathways.

\newpage

\section{Full results}
\label{appendix:allresults}

\subsection{Full information spectra on four datasets}
\label{appendix:d1}

This section provides the full information spectra, including all four PID components ($R, U_1, U_2, S$) and 
the corresponding accuracy for all 26 models across the four evaluated datasets. 
These figures supplement our main analysis in Section~\ref{res}, where we focused on 
the two most discriminative components, synergy $S$ and language uniqueness $U_2$.

\begin{figure}[h]
    \centering
    \includegraphics[width=0.95\textwidth]{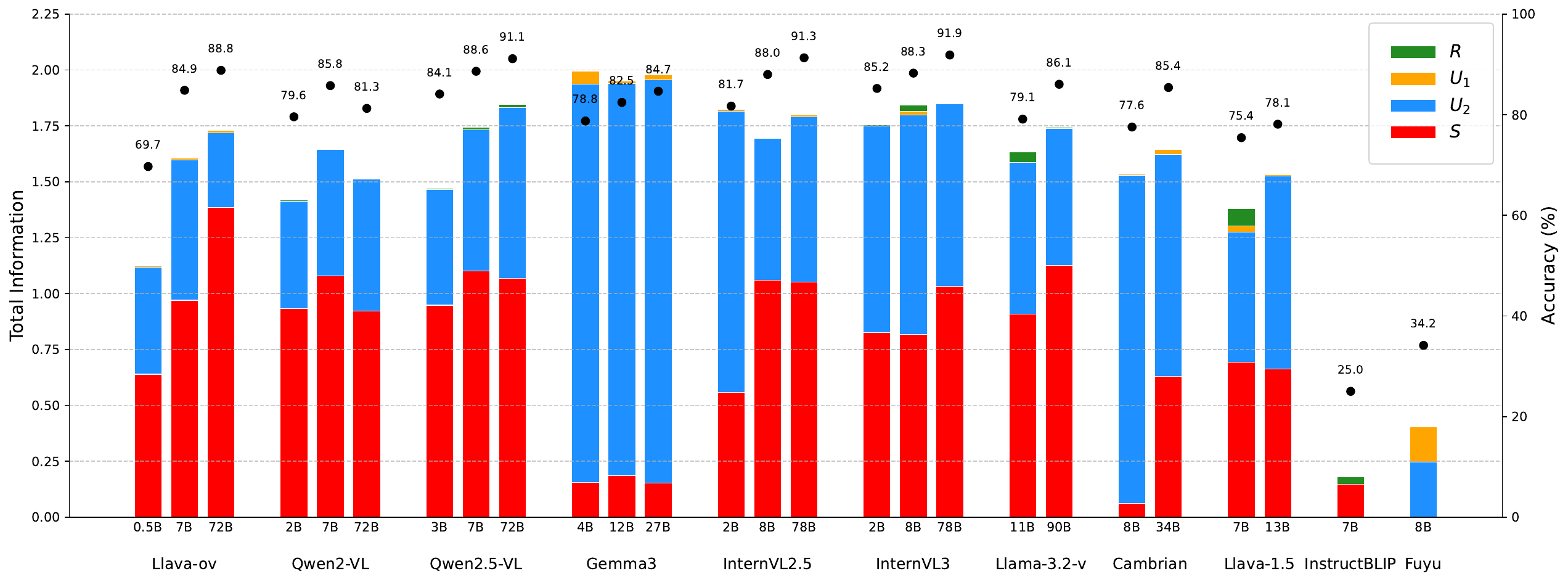}
    \caption{Full information spectrum and accuracy on the MMBench dataset. This figure provides 
    detailed evidence for the synergy-driven regime. It visually confirms the existence of two primary 
    model strategies: fusion-centric families (e.g., LLaVA-ov, Qwen2.5-VL, InternVL3) show a large share of 
    synergy $S$, while language-centric families (e.g., Gemma3) are dominated by language uniqueness $U_2$. 
    Furthermore, the plot illustrates that scaling models tends to increase synergy $S$ within many families.}
    \label{appendix:mm}
\end{figure}

\begin{figure}[h]
    \centering
    \includegraphics[width=0.95\textwidth]{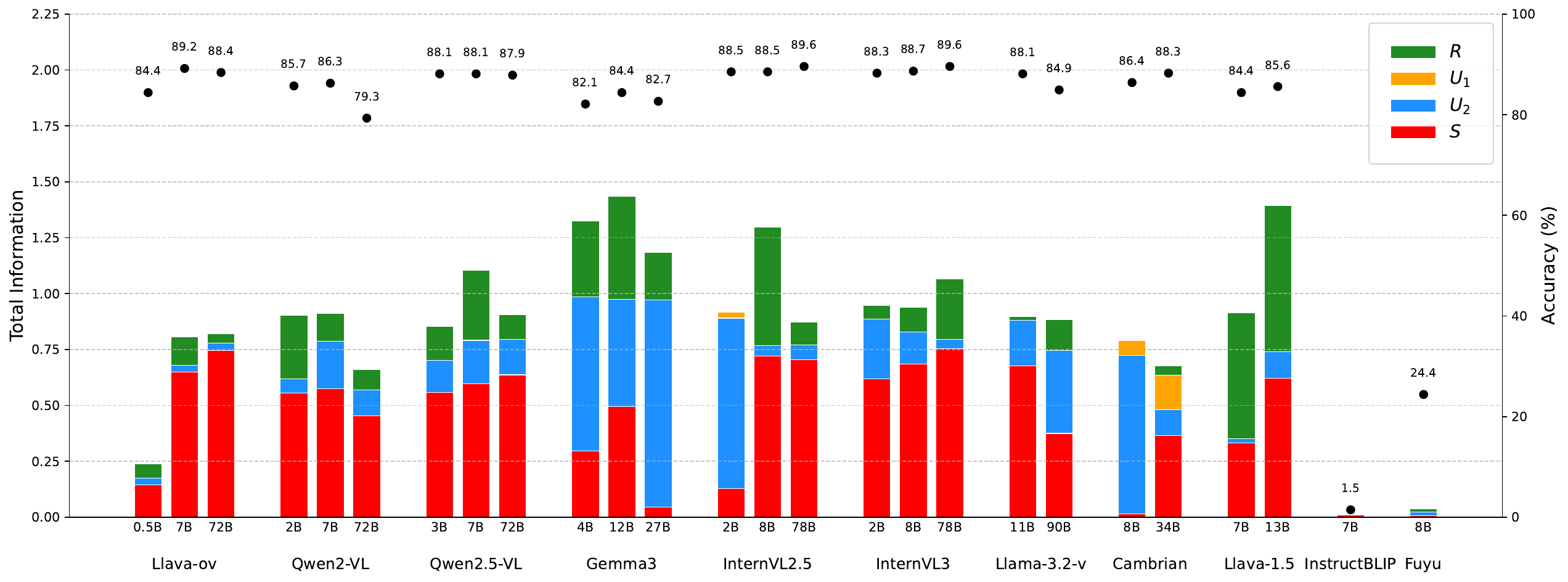}
    \caption{Full information spectrum and accuracy on the POPE dataset. As a synergy-driven task, 
    POPE shows that synergy $S$ is a key component for many high-performing models. 
    Uniquely, this dataset also elicits significant redundancy $R$ (green bars). 
    This is likely because the task uses simple binary questions about common objects (from COCO \citep{lin2014microsoft}). 
    In this context, both the visual modality (seeing the object) and the 
    language modality (understanding the object's name) can independently 
    confirm the object's presence, leading to high informational overlap.}
    \label{appendix:pope}
\end{figure}

\clearpage

\begin{figure}[h]
    \centering
    \includegraphics[width=0.95\textwidth]{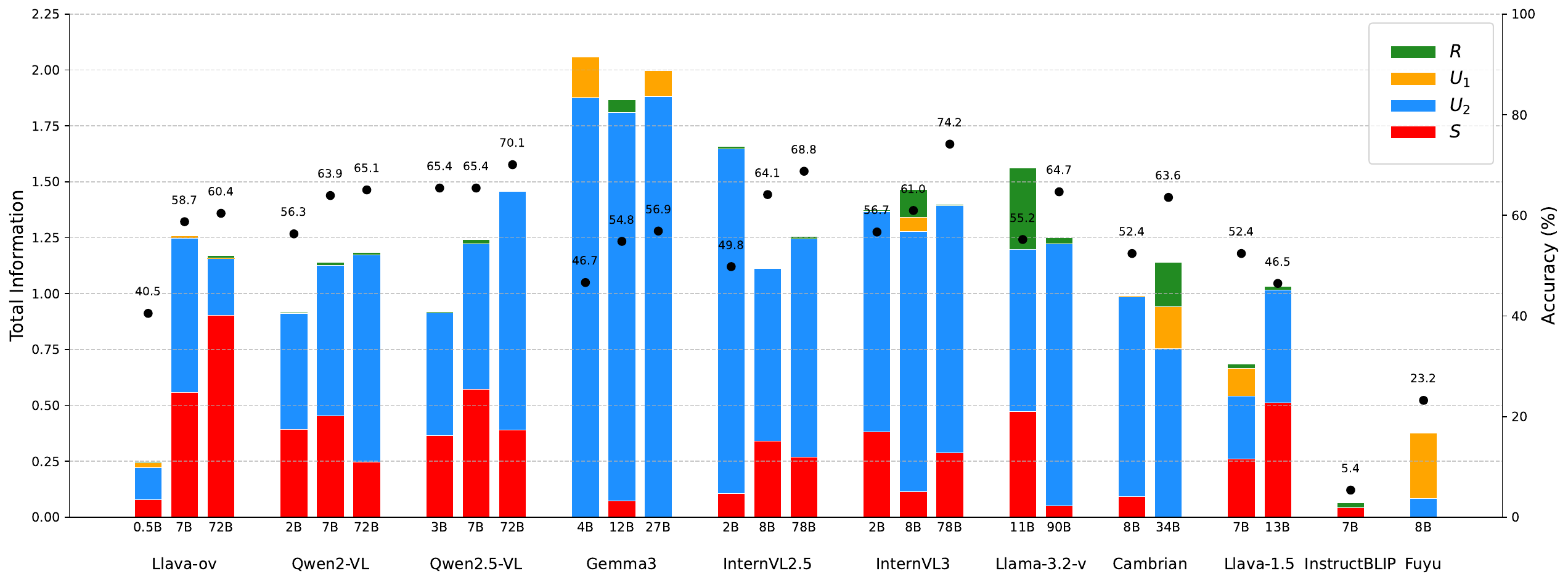}
    \caption{Full information spectrum and accuracy on Reefknot. 
    This plot exemplifies the knowledge-driven regime. 
    Language uniqueness $U_2$ is the overwhelmingly dominant 
    information component for nearly all models, demonstrating that 
    performance is constrained by language-side priors. 
    Even in this environment, the fundamental strategies of model families persist: 
    fusion-centric models (e.g., Llama-3.2-v) still generate noticeably more synergy $S$ than their 
    language-centric counterparts (e.g., Gemma3 and Cambrian).}
    \label{appendix:reefknot}
\end{figure}

\begin{figure}[h]
    \centering
    \includegraphics[width=0.95\textwidth]{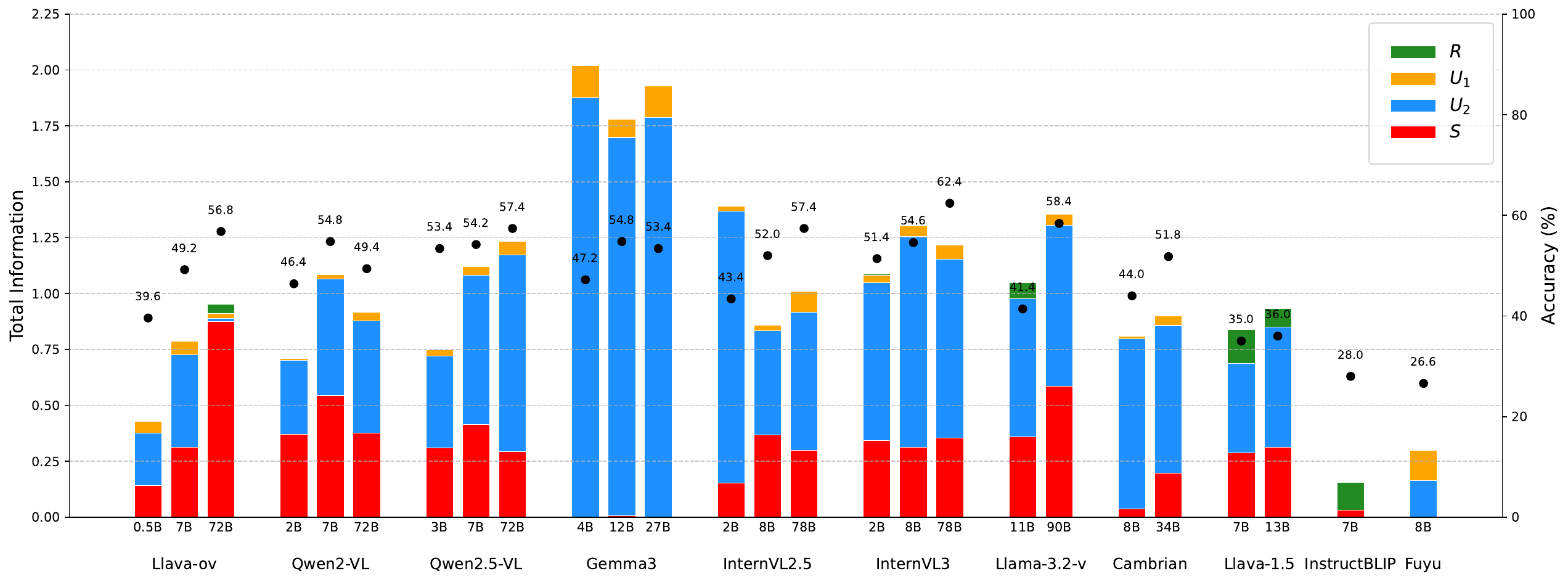}
    \caption{
    Full information spectrum and accuracy on PMC-VQA. 
    As a specialized, domain-specific task, this dataset clearly exemplifies the knowledge-driven regime. 
    Similar to Reefknot, performance is heavily dominated by language uniqueness $U_2$, 
    confirming that models must rely on internal, language-based knowledge for specialized topics. 
    This result strongly supports our finding that for domain-specific tasks, 
    performance is fundamentally limited by a model's internal, 
    language-based knowledge. The negligible synergy $S$ in many top models highlights 
    the challenge of multimodal fusion when highly specific domain knowledge is required.}
    \label{appendix:pmc}
\end{figure}

\clearpage

\subsection{Full layer-wise results on MMBench and PMC-VQA}
\label{appendix:d2}

This section provides the full, layer-by-layer information spectra for the 
representative models discussed in our main analysis. These plots show the values for 
all four PID components ($R, U_1, U_2, S$) across transformer blocks for each model 
on both MMBench and PMC-VQA. They provide the detailed evidence for the three-phase layer-wise dynamics
shared among different LVLMs.

\begin{figure}[htbp!]
    \centering
    \includegraphics[width=0.9\textwidth]{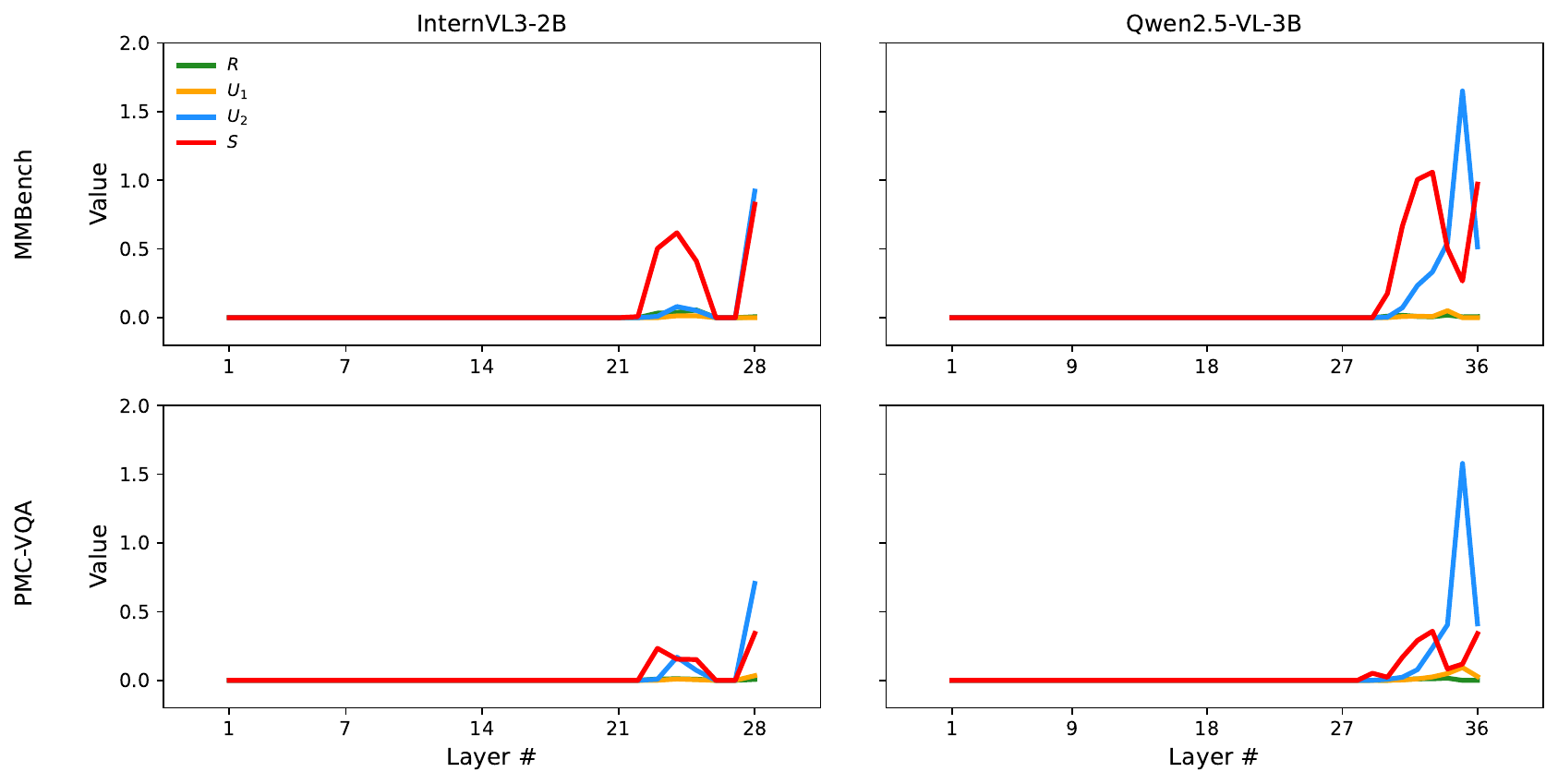}
    \caption{Layer-wise PID dynamics for InternVL3-2B and Qwen2.5-VL-3B. The plot for Qwen2.5-VL-3B clearly illustrates the standard 
    three-phase reasoning process: after information emerges, the later layers 
    show a phase of \textbf{representation} building (rising $U_2$), which 
    culminates in a decisive \textbf{fusion event} at the final 
    layer (a spike in $S$ and a drop in $U_2$). 
    In contrast, \textbf{InternVL3-2B} lacks the final fusion event; 
    its language uniqueness $U_2$ continues to rise into the final layer without the characteristic drop. 
    This may be due to its relatively smaller and shallower LLM, which might not have the 
    capacity for the distinct final-layer fusion seen in other models.}
    \label{appendix:layer1}
\end{figure}

\begin{figure}[htbp!]
    \centering
    \includegraphics[width=0.9\textwidth]{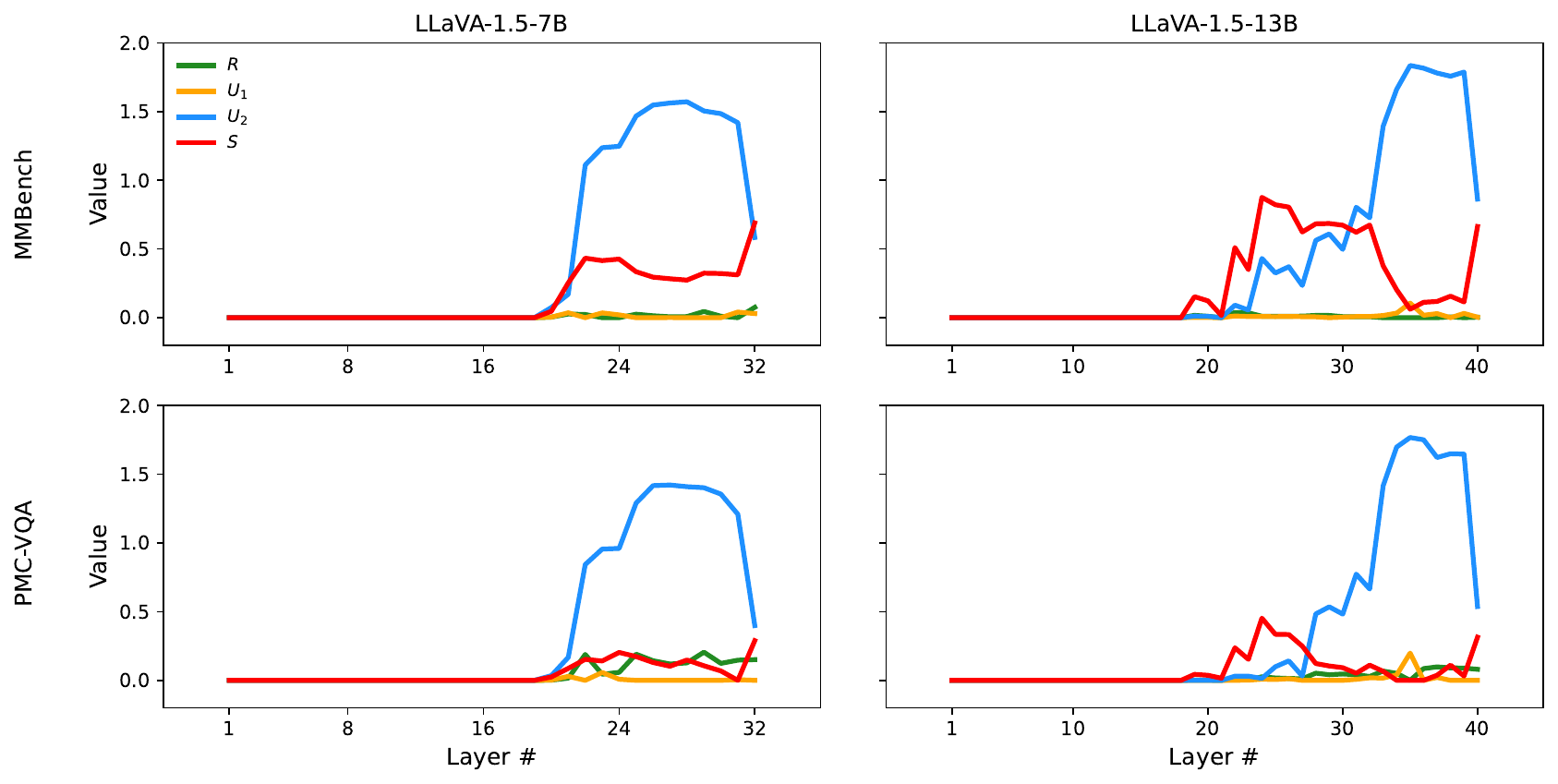}
    \caption{Layer-wise PID dynamics for LLaVA-1.5-7B and LLaVA-1.5-13B. 
    The LLaVA-1.5 family also exhibits the three-phase reasoning process, 
    though with slightly different characteristics. Information emerges in the middle layers of the network. 
    The representation building phase is distinct, with language uniqueness $U_2$ rising to a 
    high plateau while synergy $S$ forms a noticeable ``hump,'' suggesting an ongoing fusion 
    process prior to the final layer. 
    The process concludes with the characteristic fusion event, marked by a drop in $U_2$ and a final 
    spike in $S$ at the output layer. Unlike the other models, LLaVA-1.5 shows some minor, 
    non-zero activity for redundancy $R$ and vision uniqueness $U_1$ in its later layers.}
    \label{appendix:layer2}
\end{figure}

\clearpage

\subsection{Full learning-dynamics results on MMBench and PMC-VQA}
\label{appendix:d3}

This section provides the full information spectra traced across the eight training 
checkpoints of LLaVA-1.5 (7B and 13B). These plots show the values for all 
four PID components ($R, U_1, U_2, S$) on both MMBench and PMC-VQA. They provide the detailed 
evidence for the learning dynamics and scale-dependent effects identified in Finding~6.

\begin{figure}[htbp!]
    \centering
    \includegraphics[width=0.85\textwidth]{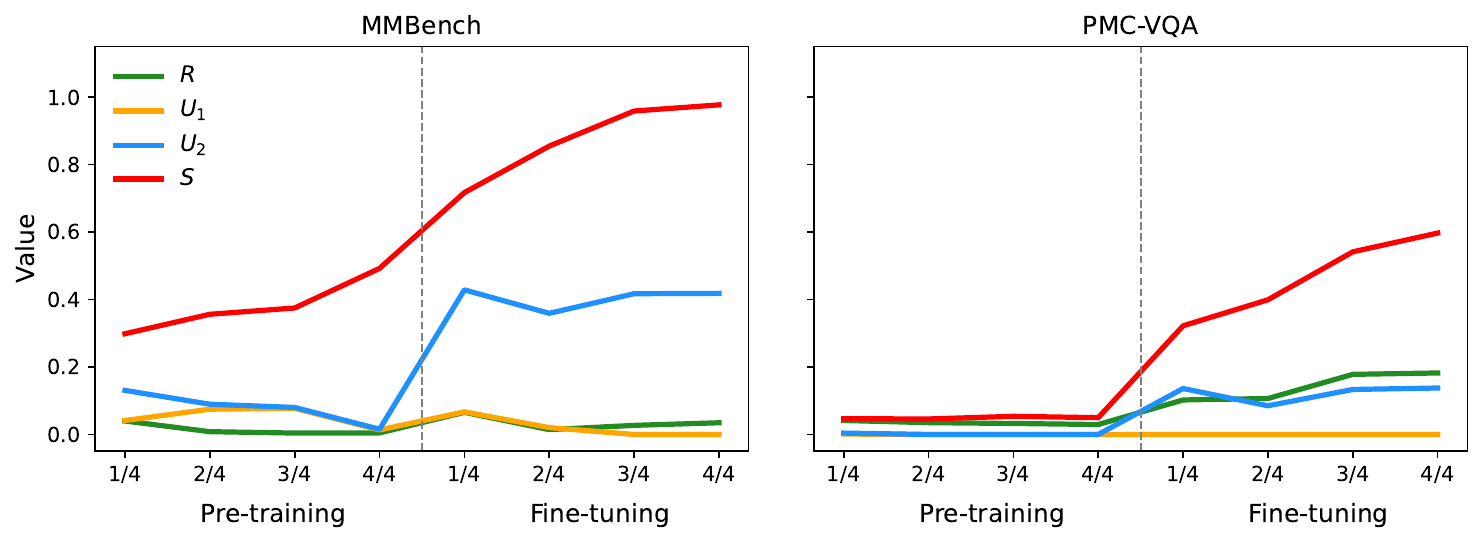}
    \caption{Learning dynamics of LLaVA-1.5-7B. This figure provides the evidence 
    for the first part of Finding~6, showing how smaller models develop 
    fusion. During Stage 1 (Pre-training), all PID components are negligible. 
    Upon commencing Stage 2 (Visual Instruction Tuning), there is a dramatic and 
    sustained increase in synergy $S$, which becomes the dominant information component 
    by the end of training. Language uniqueness $U_2$ also increases but to a 
    much lesser extent, confirming that the 7B model primarily prioritizes developing synergistic inference.}
    \label{appendix:train1}
\end{figure}

\begin{figure}[htbp!]
    \centering
    \includegraphics[width=0.85\textwidth]{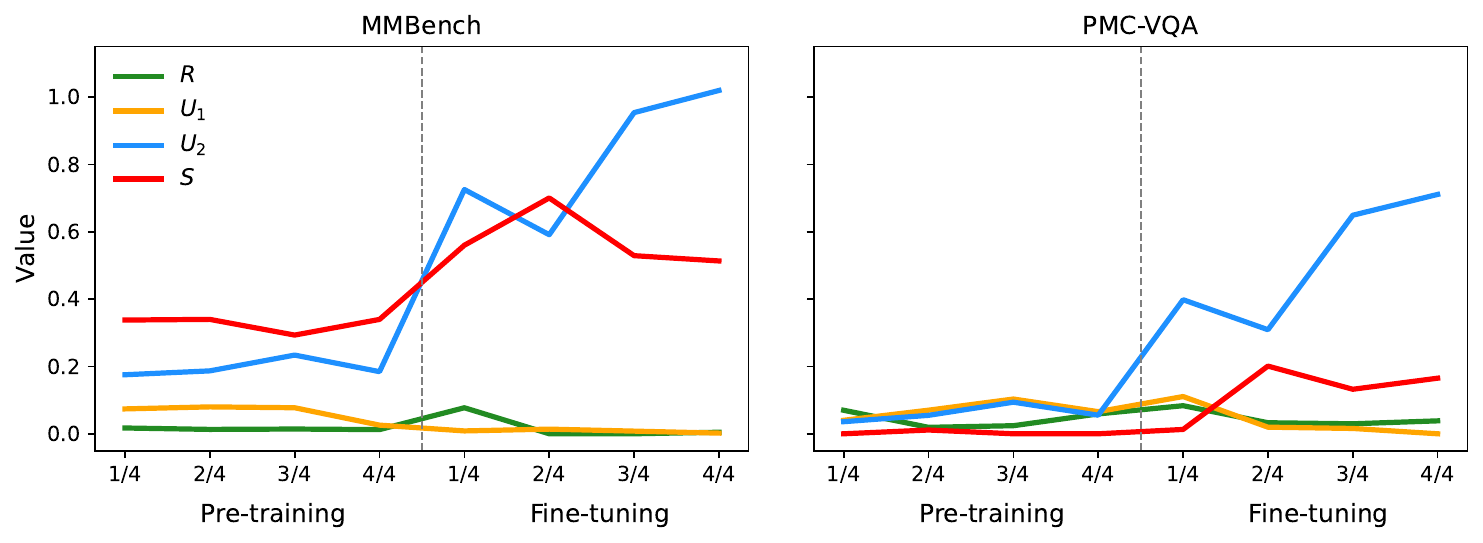}
    \caption{Learning dynamics of LLaVA-1.5-13B. In contrast to the 7B model, this figure illustrates 
    the second part of Finding~6. While PID values are 
    also flat during Stage 1, the larger 13B model exhibits a massive and 
    continuous increase in language uniqueness $U_2$ during Stage 2, becoming by 
    far the dominant information component. Although synergy $S$ also grows, its increase is 
    less pronounced than that of $U_2$, demonstrating that larger models prioritize 
    enhancing their language-side priors during visual instruction tuning.}
    \label{appendix:train2}
\end{figure}

\end{document}